\documentclass{article} 
\usepackage{iclr2025_conference,times}

\usepackage{amsmath,amsfonts,bm}

\def\eqref#1{equation~\ref{#1}}

\def\1{\bm{1}}

\DeclareMathAlphabet{\mathsfit}{\encodingdefault}{\sfdefault}{m}{sl}
\SetMathAlphabet{\mathsfit}{bold}{\encodingdefault}{\sfdefault}{bx}{n}

\usepackage{tabularx} 
\usepackage{booktabs}
\usepackage{times}
\usepackage{latexsym}
\usepackage{tabularx}
\usepackage{svg}
\usepackage{tikz}
\usepackage{subcaption}
\usetikzlibrary{shapes.geometric, arrows, positioning, arrows.meta}
\usepackage{mdframed}
\usepackage[T1]{fontenc}
\usepackage[utf8]{inputenc}
\usepackage{microtype}
\usepackage{inconsolata}
\usepackage{graphicx} 
\usepackage{float}
\usepackage{url} 
\usepackage{amsthm}
\usepackage{natbib}

\usepackage{listings}
\usepackage{xcolor}
\definecolor{ForestGreen}{RGB}{34,139,34}

\usepackage{enumitem}
\usepackage{hyperref}

\title{Epistemic Integrity in Large Language Models}

\author{\hspace{-0.8mm}
\textbf{Bijean Ghafouri\thanks{Equal contribution.}\hspace{1.5mm}\textsuperscript{1}}
\And
 \textbf{Shahrad Mohammadzadeh\footnotemark[1]\hspace{1.5mm}\textsuperscript{2,10}}
 \And
 \textbf{James Zhou\textsuperscript{3}}
 \AND
 \textbf{Pratheeksha Nair\textsuperscript{2,10}}
 \And
 \textbf{Jacob-Junqi Tian\textsuperscript{4,10}}
 \And
 \textbf{Hikaru Tsujimura\textsuperscript{5,6}}
 \AND
 \textbf{Mayank Goel\textsuperscript{7}}
 \And
 \textbf{Sukanya Krishna\textsuperscript{8}}
 \And
 \textbf{Reihaneh Rabbany\textsuperscript{2,10}}
 \AND
 \textbf{Jean-François Godbout\textsuperscript{9,10}}
 \And
 \textbf{Kellin Pelrine\textsuperscript{2,10}}
\AND
\normalfont
 \textsuperscript{1}University of Southern California\hspace{0.8mm}
 \textsuperscript{2}McGill University\hspace{0.8mm}
 \textsuperscript{3}UC Berkeley\hspace{0.8mm}
 \textsuperscript{4}Vector Institute\\
 \textsuperscript{5}Cardiff University\hspace{0.8mm}
 \textsuperscript{6}University College London\hspace{0.8mm}
 \textsuperscript{7}IIIT Hyderabad\hspace{0.8mm}
 \textsuperscript{8}Harvard University\hspace{0.8mm}\\
 \textsuperscript{9}Université de Montréal\hspace{0.8mm}
 \textsuperscript{10}Mila
\\
 \small{
   \textbf{Correspondence:} \href{kellin.pelrine@mila.quebec}{kellin.pelrine@mila.quebec}
 }
}

\iclrfinalcopy 
\begin{document}

\maketitle
\begin{center}
\end{center}

\begin{abstract}
    Large language models are increasingly relied upon as sources of information, but their propensity for generating false or misleading statements with high confidence poses risks for users and society. In this paper, we confront the critical problem of epistemic miscalibration — where a model's linguistic assertiveness fails to reflect its true internal certainty. We introduce a new human-labeled dataset and a novel method for measuring the linguistic assertiveness of Large Language Models (LLMs) which cuts MSE by over 50\% relative to previous benchmarks. Validated across multiple datasets, our method reveals a stark misalignment between how confidently models linguistically present information and their actual accuracy. Further human evaluations confirm the severity of this miscalibration. This evidence underscores the urgent risk of the overstated certainty LLMs hold which may mislead users on a massive scale. Our framework provides a crucial step forward in diagnosing this miscalibration, offering a path towards correcting it and more trustworthy AI across domains.
\end{abstract}

\section{Introduction}
Large Language Models (LLMs) have markedly transformed how humans seek and consume information, becoming integral across diverse fields such as public health \citep{ali2023using}, coding \citep{zambrano2023from}, and education \citep{whalen2023chatgpt}. Despite their growing influence, LLMs are not without shortcomings. One notable issue is the potential for generating responses that, while convincing, may be inaccurate or nonsensical---a long-standing phenomenon often referred to as ``hallucinations'' \citep{jo2023promise, huang2023survey, zhou2024larger}. This raises concerns about the reliability and trustworthiness of these models.

A critical aspect of trustworthiness in LLMs is \textit{epistemic calibration}, which represents the alignment between a model's internal confidence in its outputs and the way it expresses that confidence through natural language. Misalignment between internal certainty and external expression can lead to users being misled by overconfident or underconfident statements, posing significant risks in high-stakes domains such as legal advice, medical diagnosis, and misinformation detection. While of great normative concern, how LLMs express linguistic uncertainty has received relatively little attention to date \citep{sileo2023probing, belem2024perceptions}. 

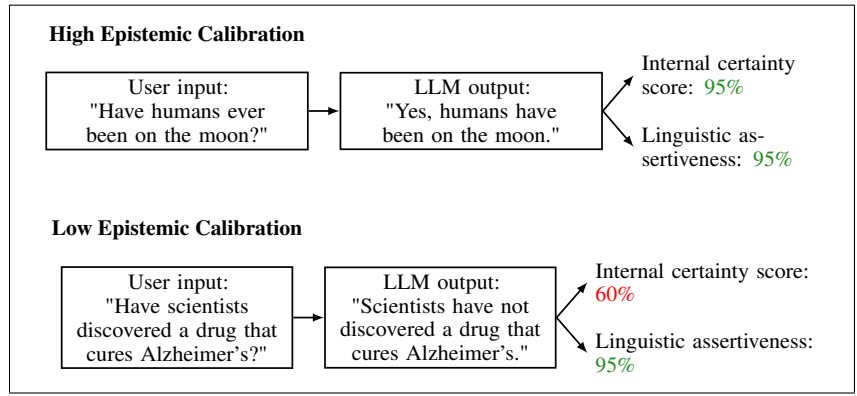
\begin{figure}[ht]
\label{figure:epistemic_example}
\centering
\begin{mdframed}[style=MyFrame,nobreak=true,align=center,userdefinedwidth=32em] 
\resizebox{\textwidth}{!}{
\begin{tikzpicture}[node distance=0.3cm and 0.5cm, auto, thick, >=Stealth] 
    \node (title1) [align=center] {\textbf{High Epistemic Calibration}};
    \node (input1) [rectangle, draw, text width=4cm, below=of title1, minimum height=0.7cm, align=center] {User input: \\
    "Have humans ever been on the moon?"};
    \node (output1) [rectangle, draw, right=of input1, text width=4cm, minimum height=1.1cm, align=center] {LLM output: \\
    "Yes, humans have been on the moon."};
    \node (internal1) [above right=of output1, text width=2.8cm, yshift=-0.8cm] {Internal certainty score: \textcolor{ForestGreen}{95\%}};
    \node (external1) [below right=of output1, text width=2.8cm, yshift=0.8cm] {Linguistic assertiveness: \textcolor{ForestGreen}{95\%}};

    \node (title2) [align=center, below=1cm of input1] {\textbf{Low Epistemic Calibration}};
    \node (input2) [rectangle, draw, text width=3.5cm, below=of title2, minimum height=1.2cm, align=center] {User input:\\
    "Have scientists discovered a drug that cures Alzheimer's?"};
    \node (output2) [rectangle, draw, right=of input2, text width=3.5cm, minimum height=1.2cm, align=center] {LLM output:\\
    "Scientists have not discovered a drug that cures Alzheimer's."};
    \node (internal2) [above right=of output2, text width=3.5cm, yshift=-1.0cm] {Internal certainty score: \textcolor{red}{60\%}};
    \node (external2) [below right=of output2, text width=3.5cm, yshift=1.0cm] {Linguistic assertiveness: \textcolor{ForestGreen}{95\%}};

    \draw[-latex] (input1.east) -- (output1.west); 
    \draw[-latex] (output1.east) -- (internal1.west); 
    \draw[-latex] (output1.east) -- (external1.west); 
    
    \draw[-latex] (input2.east) -- (output2.west); 
    \draw[-latex] (output2.east) -- (internal2.west); 
    \draw[-latex] (output2.east) -- (external2.west); 
\end{tikzpicture}
}
\end{mdframed}
\caption{\small This figure illustrates two examples with varying levels of epistemic calibration in LLM outputs. The one below is poorly calibrated. For each output, we calculate two certainty scores: internal certainty and external certainty (linguistic assertiveness). The internal certainty is computed using the method outlined by \citet{pelrine2023reliable}. To assess linguistic assertiveness, we develop a custom model, which we validate using human ratings collected through a survey.}
\label{fig:llm_calibration_comparison}
\end{figure}

Figures \ref{fig:llm_calibration_comparison} and \ref{fig:full_width_flowchart} illustrate the issue of epistemic calibration providing insights into the operation of certainty in the context of human interactions with LLMs. We highlight the following key points in these figures: 
\begin{itemize}[leftmargin=10pt,topsep=-2pt,noitemsep]
\item \textbf{Distinct Roles of Certainty:} Internal certainty and linguistic assertiveness have distinct functions within LLM interactions that shape individual beliefs.
    \item \textbf{Human access to LLM certainty:} Linguistic assertiveness holds a critical role as the primary form of certainty available to users. Unlike internal certainty, which remains hidden within the model's computational processes, linguistic assertiveness is directly perceivable and influences how users interpret the model's outputs.
    \item \textbf{Beyond Content:} Users retrieve more than just the content from an LLM's output. The style and assertiveness of the language used also play a significant role by shaping perceptions through the communication of certainty. This interaction between the model's output and its linguistic assertiveness is crucial for understanding the full impact on individual perceptions. 
\end{itemize}

\mdfdefinestyle{MyFrame}{
    linecolor=black, 
}

Several studies have explored the calibration of internal confidence in LLMs. For instance, \citet{zhang2024r} examine confidence calibration, proposing techniques to reduce hallucinations and enhance the model's ability to answer known questions while avoiding unknown ones. However, they overlook the role of \textit{linguistic assertiveness} and how external certainty can still lead to epistemic miscalibration even if internal confidence is addressed. Similarly, \citet{ren2023investigating} focus on factual knowledge and LLM behavior before and after retrieval-augmented generation (RAG). While they investigate internal confidence, they fail to frame miscalibration as an end-to-end issue involving both internal certainty and linguistic assertiveness, therefore ignoring the interplay between model predictions and how confidence is expressed linguistically.
Since this study was first conducted, emerging work has sharpened the theoretical distinction between internal confidence and linguistic assertiveness, emphasizing the potential for "epistemic mismatch"—where LLMs express unwarranted certainty in their generated outputs despite low internal confidence. Recent empirical studies have shown that models tend to default to confident, assertive language even when their token-level certainty or reasoning reliability is low (\cite{leng2025tamingoverconfidencellmsreward}; \cite{zhou2025calibratingllmconfidencesemantic}; \cite{wen2024from}). This growing literature underscores the need for methods that can measure and calibrate linguistic certainty directly, particularly in high-stakes domains.

Studies have attempted to bridge the gap between internal confidence and linguistic assertiveness but still face considerable limitations. Although \citet{mielke2022reducing} explore epistemic calibration, their study uses a limited scoring scale to measure both assertiveness and confidence, which restricts the continuous assessment of LLM output. Their approach also relies on a narrow range of datasets, thereby limiting its applicability across domains. \citet{zhou2024relying} address miscalibration using epistemic markers, but their method lacks real domain grounding and fails to consider the complexity of natural language. 

Framing these developments within the broader context of trust and epistemic communication, concurrent work by \citet{yona2024can} provides strong empirical evidence that large language models often express undue linguistic confidence—answering decisively even when internally uncertain. This misalignment, which they formalize as faithful response uncertainty, reveals that prompting alone is insufficient to ensure models communicate their uncertainty reliably. The resulting ``epistemic mismatch'' not only hinders trustworthiness but also exacerbates overreliance in knowledge-seeking tasks. Our work contributes to this conversation by offering a model-driven approach to quantifying assertiveness, enabling more precise detection of cases where linguistic style diverges from underlying confidence.

This review of existing work on LLM calibration and confidence reveals several limitations that our research aims to address:

\begin{itemize}[leftmargin=10pt,topsep=2pt,noitemsep]
    \item \textbf{Few Integrated Approaches}: Previous studies typically address either internal certainty or linguistic assertiveness but rarely both simultaneously \citep{jiang2021how}. There is a need for comprehensive frameworks that integrate these aspects to ensure LLMs communicate accurately and responsibly.

    \item \textbf{Inadequate Assertiveness Measurement}: Existing methods for measuring linguistic assertiveness often rely heavily on lexicon-based approaches \citep{pei2021measuring} or subjective perceptions without adequate validation \citep{steyvers2024calibration}. These methods can lack contextual depth and fail to generalize across diverse domains.

    \item \textbf{Limited High-Stakes Evaluation}: Although some studies explore epistemic calibration, they mainly cover a narrow range of topics and employ low-resolution measures of assertiveness, limiting their applicability in critical domains such as misinformation detection \citep{mielke2022reducing}.
\end{itemize}

To address these gaps, our paper provides:

    \begin{itemize}[leftmargin=10pt,topsep=2pt,noitemsep] \item \textbf{A New Assertiveness Detection Model}: We train a new model to detect linguistic assertiveness, using a composite of five diverse datasets. Our approach improves the accuracy relative to previous approaches by incorporating contextual nuance and aligning more closely with human perceptions. We also address limitations in previous methods in generalizably measuring assertiveness across domains.
    
    \item \textbf{Empirical Evidence of Epistemic Miscalibration}: Our work provides a comprehensive comparison between internal certainty and linguistic assertiveness, documenting instances of miscalibration in different contexts. Our experiments reveal that LLMs frequently generate highly assertive explanations despite low internal certainty, which can mislead users.

    \item \textbf{Validation with Human Perception}: We conduct comprehensive surveys assessing human perceptions of LLMs' linguistic assertiveness. Our results confirm that there is a misalignment between computational measures and subjective human perceptions of language, highlighting the need for more robust linguistic calibration of LLMs.
\end{itemize}

This study also presents a human-centered approach for developing a robust assertiveness scoring method. To ensure reliability, we train and compare several models to identify the best estimator for assertiveness based on accuracy and transferability using a new multi-domain dataset hand-coded to measure certainty. After selecting the top-performing model, we validate the results using a different human-surveyed dataset \citep{wang2017liar}, which was coded independently by different individuals. This comprehensive methodology enables us to thoroughly assess both the objective and subjective aspects of assertiveness in language model explanations. 

Our findings reveal that when the model has low internal certainty, it generates explanations that are significantly over-assertive, meaning the language used implies a higher degree of certainty than is warranted by the model's actual confidence or accuracy. This miscalibration could lead users to misconstrue the model's judgments as more reliable than they actually are. More precisely, our results confirm a strong correlation between GPT-4o's assertiveness scores and human perceptions of assertiveness, but a weak correlation between human perceptions and internal certainty, and an even weaker relationship between GPT-4o model assertiveness and internal certainty. Together, these results provide strong evidence of epistemic miscalibration in LLMs.\footnote{For the code and datasets used, refer to our GitHub repository at: \url{https://github.com/ComplexData-MILA/epistemic-integrity}.}


\section{Conceptually understanding certainty and assertiveness in natural language}

To better understand the challenges of epistemic calibration, it is essential to define the concepts of certainty and assertiveness in natural language communication. These foundational notions explain how information is conveyed by LLMs and interpreted by users.

\subsection{Certainty}
Effective communication hinges on the accurate conveyance of certainty, enabling individuals and systems to assess the reliability of information. In human communication, speakers use linguistic cues to express their confidence levels, which listeners interpret to form judgments about the truthfulness and credibility of statements \citep{budescu1985consistency, clarke1992ratings}. Similarly, for LLMs, effectively conveying certainty is crucial to ensure users can trust and interpret the provided information accurately. In this section, we decompose \textit{certainty} into two key concepts: \textbf{Internal Certainty} and \textbf{External Certainty}. Later, we argue in Section \ref{sec:epistemic_calibration} that misalignment between these two dimensions requires \textbf{Epistemic Calibration}.

\subsubsection{Internal Certainty}
Internal certainty, also referred to as model confidence, represents the probability that an LLM assigns to a particular output based on its internal computations and parametric knowledge from its training data. In tasks such as question-answering, internal certainty is often represented by the probability the model assigns to its selected response compared to alternative answers \citep{jiang2021how, hendrycks2021massive}. For instance, when generating an answer, the model evaluates the likelihood of various possible responses, expressing its level of confidence in the chosen output.

\paragraph{Internal Confidence Estimation}
\citet{hendrycks2021massive} introduce baseline methods to detect misclassifications by examining model confidence. One common approach is token-level analysis, which examines the probability assigned to individual tokens to generate fine-grained estimates of uncertainty \citep{jiang2021how, kuhn2023semantic, duan2024shifting}. Another method involves assessing the levels of variability across multiple outputs generated from the same input, where greater variability suggests higher uncertainty \citep{xiong2024can}. \citep{shrivastava2023llamas} offer an additional technique by employing external classifiers trained on both input data and the model's internal representations to predict uncertainty, which provides a more comprehensive assessment of the model's internal state.

Recent research has also focused on making internal confidence more interpretable for users. Some approaches express internal certainty numerically (e.g., ``80\% confidence'') \citep{lin2022teaching, xiong2024can}, while others adopt qualitative expressions (e.g., ``I am not confident in the answer'') \citep{mielke2022reducing, zhou2023navigating}. These strategies aim to make internal confidence scores more accessible by incorporating them directly into the model's natural language outputs.

\paragraph{Challenges in Confidence Alignment}
As research on estimating the internal confidence of LLMs has increased, scholars have started to focus on model calibration---i.e., the alignment between predicted probabilities and actual correctness. \citet{desai2020calibration} find that pre-trained transformers such as BERT often exhibit poor calibration out-of-the-box, where their confidence estimates fail to correspond to actual correctness. \citet{jiang2021how} also explore methods to improve model calibration, such as temperature scaling, which adjusts predicted probabilities to better match actual outcomes. However, most calibration methods focus on probabilistic outputs (internal confidence) without addressing how certainty is expressed through language generation. The same miscalibration is found by \citet{si2022prompting}, who demonstrate that models like GPT-3 frequently produce overconfident responses even when incorrect.

However, we note that despite these known limits, internal confidence scores remain largely inaccessible to users today. Without these scores, it becomes essential for models to effectively communicate uncertainty through external means---such as linguistic cues---to ensure users correctly interpret the model's output. 

\subsubsection{External Certainty}

External certainty refers to the level of confidence conveyed through the textual generation of an LLM, as interpreted by an external observer. This type of certainty reflects how assertive, definitive, or unambiguous the model's output appears, regardless of the underlying internal confidence scores generated during the prediction process \citep{mielke2022reducing}. 

A critical component of external certainty is \textbf{Linguistic Assertiveness}, which involves the use of linguistic markers—such as modal verbs, adverbs, and other cues—that signal varying degrees of confidence or uncertainty. For example, statements like ``it is certain that'' and ``there is a possibility that'' differ significantly in their level of assertiveness. We note that such differences could influence how users perceive the reliability of the information presented.

Human communication tends to favor the expression of uncertainty through linguistic means rather than numerical values \citep{erev1990verbal, wallsten1993preferences}. Research on how individuals interpret verbal cues of uncertainty reveals that people map linguistic expressions of confidence onto numerical values in different ways, depending on the context, expertise, and domain in question \citep{windschitl1996measuring, karelitz2004you}. Although there is some variability in individual interpretations, population-level studies show consistent patterns in how verbal expressions of uncertainty correspond to probabilistic estimates \citep{budescu1985consistency, clarke1992ratings}.

\paragraph{Linguistic Assertiveness in LLM Outputs}
\citet{mielke2022reducing} examine how both humans and LLMs interpret and generate expressions of uncertainty. They find that LLMs such as GPT-4 can map uncertainty expressions to numerical equivalents similarly to humans, though they are more prone to biases rooted in their training data. However, most prior research focuses on how LLMs interpret uncertainty expressions rather than how they generate assertive or uncertain language aligned with their internal confidence scores.
These developments also align with broader insights from reinforcement learning with human feedback (RLHF), where model behaviors that mirror human-like rhetorical confidence can be implicitly rewarded, regardless of factual accuracy (\cite{leng2025tamingoverconfidencellmsreward}). Studies have also found that prompt-level variations can influence perceived certainty (\cite{zhou2025calibratingllmconfidencesemantic}), and that task format and cognitive complexity can exacerbate miscalibration (\cite{chhikara2025mindconfidencegapoverconfidence}). These findings further validate the need for domain-agnostic tools that assess and align assertiveness with actual model certainty, motivating our focus on assertiveness quantification.

This gap between a model's internal certainty and its external linguistic expressions is particularly important to address, as users often rely on the model's language to gauge its confidence. Our work seeks to bridge this gap by analyzing how linguistic assertiveness in model outputs correlates with internal certainty, thereby contributing to the broader goal of epistemic calibration.

\subsection{Epistemic Calibration}
\label{sec:epistemic_calibration}
Epistemic Calibration is the process of aligning a model's expressed confidence, conveyed through linguistic assertiveness, with its actual reliability or correctness. Achieving this alignment requires that the model's linguistic expressions match the probabilistic confidence it has in its predictions. However, if this balance is disrupted, significant communication issues can arise. For instance, if a model uses assertive language but is internally uncertain, users may place undue trust in potentially incorrect information. Conversely, when a model is internally confident but hedges its language, users might doubt accurate information.

Figure \ref{fig:llm_calibration_comparison} presents two contrasting examples that highlight varying levels of epistemic calibration in LLM outputs. In both cases, the LLM responds to a user query, and we compute two certainty scores: Internal Certainty and External Certainty (through linguistic assertiveness). In the first example, the internal and external certainty scores are closely aligned, demonstrating high epistemic calibration. Here, the model's linguistic assertiveness accurately reflects its internal confidence, resulting in a trustworthy output. In contrast, the second example reveals low epistemic calibration. Although the internal certainty score is relatively low, the model's linguistic assertiveness remains high, indicating overconfidence in its response. In the following analysis, we compute internal certainty using the method outlined by \citet{pelrine2023reliable} and use linguistic assertiveness, derived from our custom model detailed in Section \ref{sec:methods}---which is validated against human surveyed human ratings---as a proxy for external certainty quantification.

\section{Methods}
\label{sec:methods}
\subsection{Datasets and Models}
\label{section:models}

The dataset we use for the certainty calibration scoring task is the LIAR dataset, a misinformation dataset consisting of 12,800 political statements fact-checked by PolitiFact \citep{wang2017liar}. Each statement is labeled as true or false based on PolitiFact's classification.\footnote{PolitiFact provides 6-way labels; we follow standard practice by binarizing these labels \cite{pelrine2023reliable}.} To augment this dataset, we employ OpenAI's GPT-4 to reassess the veracity of each statement, providing a foundation for the certainty calibration analysis that follows.

Recognizing the limitations of previous methods for assertiveness calibration, we compile a new dataset and train our own models to achieve more accurate results. Existing approaches, such as \citet{pei2021measuring}, rely on a BERT-based model that is constrained by input length and limited to news and scientific articles, reducing their applicability to other domains. Other methods, such as \citet{Byalyk_Nizhnik_2022}, base assertiveness scores on lexicon-derived buckets, which limits their adaptability to diverse contexts. To overcome these challenges, we curated a diverse composite dataset across multiple domains to improve scalability and transferability, consisting of 800 data points equally distributed across the following five sources:

\begin{itemize}[leftmargin=10pt,topsep=2pt,noitemsep]
    \item \textbf{Anthropic's Persuasiveness dataset} \citep{durmus2024persuasion}: Text data that compares the persuasiveness of arguments generated by humans and LLMs.
    \item \textbf{Globe and Mail (GM) Comments dataset} \citep{kolhatkar2020sfu}: User-generated comments from the Globe and Mail newspaper.
    \item \textbf{Reddit Change My View (CMV) dataset} \citep{wiegmann-etal-2022-analyzing}: User texts where persuasive arguments successfully change another user's viewpoint.
    \item \textbf{Arguments generated by LLaMA 3-8B on LIAR dataset} \citep{dubey2024llama3herdmodels}: Text responses generated by LLaMA 3-8B assessing the factuality of statements in the LIAR dataset.
    \item \textbf{Pei's assertiveness dataset} \citep{pei2021measuring}: The dataset used to train Pei's assertiveness model, focused on assertiveness in scientific communication.
\end{itemize}

The data was annotated by three expert coders and eleven additional coders following the guidelines outlined in Appendix \ref{appendix:guidance}. We randomly sample 800 data points from these five sources, and inter-coder agreement, measured by the correlation between individual coders and the average score, indicates a mean agreement centered around 0.7 (see Table \ref{tab:coder_agreement} in the Appendix).

To evaluate model performance in assertiveness calibration, we test the following models:

\begin{itemize}[leftmargin=10pt,topsep=2pt,noitemsep]
\item \textbf{Pre-existing Pei \& Jurgens SciBERT model} \citep{pei2021measuring}: an existing state-of-the-art approach to measuring linguistic certainty, using a fine-tuned SciBERT model. We use the weights provided by the authors.
    \item \textbf{Fine-tuned Pei \& Jurgens SciBERT model}: SciBERT model further fine-tuned on our dataset for assertiveness scoring.  
    \item \textbf{Random Forest Nizhnik model} \citep{Byalyk_Nizhnik_2022}: Random forest model trained on a taxonomy of epistemic markers, using individual words as features for classification.
    \item \textbf{Llama-3.2-1B-Instruct fine-tuned with LoRA} \citep{hu2021loralowrankadaptationlarge}: LLaMA model fine-tuned using LoRA \citep{vonwerra2022trl}, based on the same assertiveness scoring prompt used by human coders.
    \item \textbf{Prompted GPT-4o-2024-08-06} \citep{hello_GPT4o}: Prompted like the human coders to score assertiveness on a scale of 0 to 10.
    \item \textbf{GPT-4o-2024-08-06 fine-tuned on assertiveness dataset}: Fine-tuned using OpenAI's API, applying the same assertiveness scoring prompt.
    \item \textbf{GPT-4o-2024-08-06 fine-tuned with rounding}: Fine-tuned similarly, but with assertiveness scores rounded to one decimal place in the fine-tuning data.

\end{itemize}

We evaluate model performance using standardized mean squared error (MSE) to normalize outputs across models. The best-performing models are used in subsequent miscalibration experiments to compare assertiveness scores.\footnote{For reproducibility, the models can be trained and fine-tuned using the code on our GitHub and for GPT-4o, through OpenAI API Dashboard.}

\textbf{Followup Findings (March 2025)}: We test two more models' performance in measuring assertiveness:

\begin{itemize}[leftmargin=10pt,topsep=2pt,noitemsep]
    \item \textbf{Llama-3.2-3B-Instruct fine-tuned with LoRA with rounding} \citep{hu2021loralowrankadaptationlarge}: Fine-tuned using LoRA similar to the Llama-3.2-1B model, but with assertiveness scores rounded to one decimal place in the fine-tuning data as with GPT-4o with rounding.
    \item \textbf{Llama-3.2-11B-Vision-Instruct fine-tuned with LoRA with rounding} \citep{hu2021loralowrankadaptationlarge}: a larger Llama model trained with the same process as the 3B one.
\end{itemize}


\subsection{Computing internal certainty and linguistic assertiveness}
To estimate the internal certainty of the LLM, we use the verbalized confidence method outlined in \citet{pelrine2023reliable}. This method provides a combination of strong performance and simplicity of use compared to other state-of-the-art methods for the misinformation detection domain \citep{rivera2024combining}. Specifically, the model is instructed to provide an explanation for each misinformation classification and assign an uncertainty score. The scores are calibrated on a validation set using Platt's method \citep{platt1999probabilistic}. 
The robustness of this method is further validated in Appendix~\ref{app:certainty}.


To measure the linguistic assertiveness of LLM-generated explanations in the misinformation task, we use the best-performing model from Section \ref{section:models}. We then compare this assertiveness measure to the underlying certainty estimates obtained using the uncertainty quantification techniques described above. By analyzing the gap between the model's certainty and assertiveness, we quantify the degree of calibration in its linguistic expressions.

\section{Results}
\paragraph{Assertiveness Calibration Score}

\begin{figure}[htbp]
    \centering
    \includegraphics[width=0.7\linewidth]{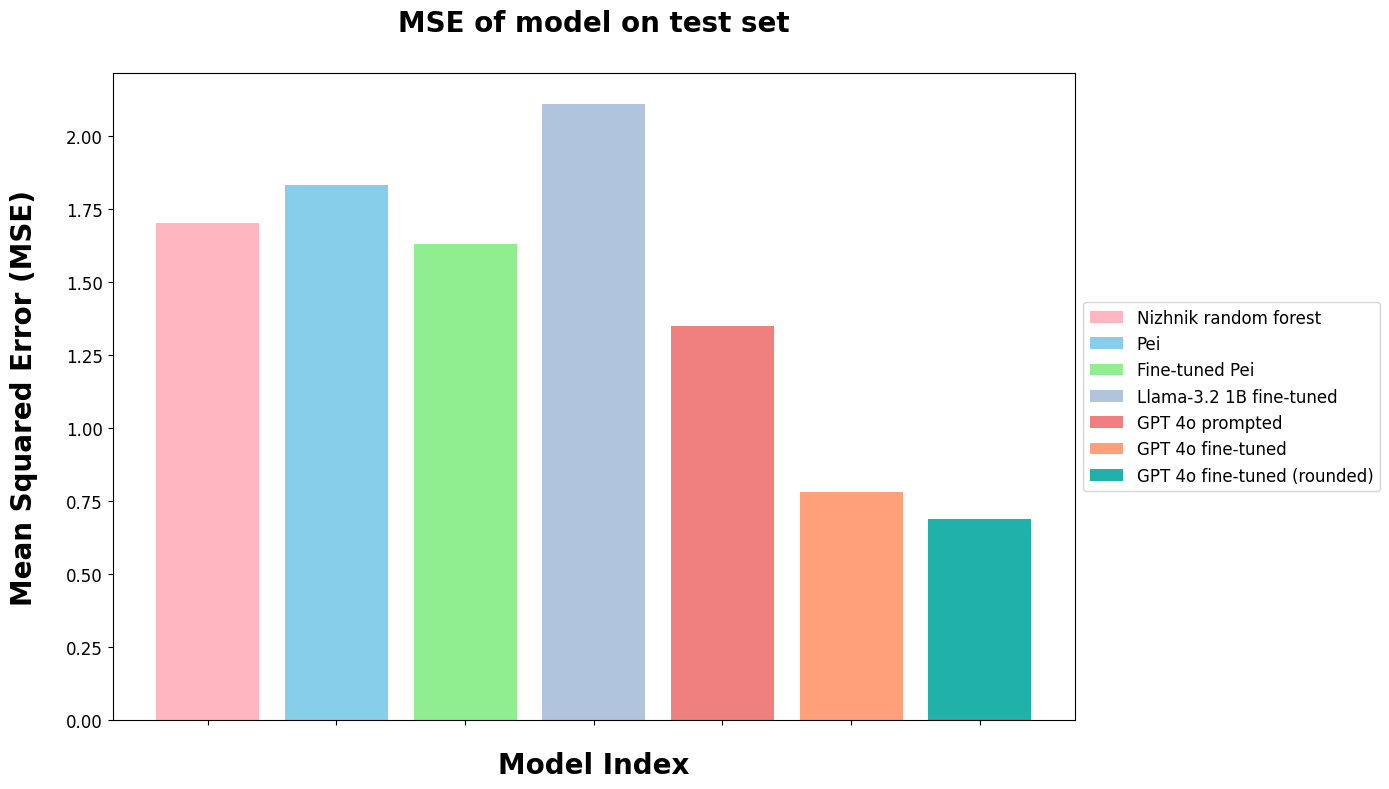}
   \caption{\small Comparison of assertiveness evaluation performance across models. The models are evaluated based on their Mean Squared Error (MSE) relative to the test set, which is a subset of our dataset. Among models tested here, the fine-tuned GPT-4o, trained with assertiveness scores rounded to one decimal point, achieved the lowest MSE, which suggests the highest accuracy in predicting assertiveness.}
    \label{Figure 1}
    \vspace{-3mm}
\end{figure}

Figure~\ref{Figure 1} is obtained from comparing the seven different methods of assertiveness quantification, evaluating on the test set of our composite dataset. We find that \textit{GPT-4o fine-tuned with rounding} (training on assertiveness scores rounded to one decimal point) achieves the highest accuracy in predicting human-annotated assertiveness scores. The margin of improvement over the approaches from the literature is very large, cutting Mean Squared Error (MSE) by more than half. To validate the transferability of these results across different domains, we conduct an ablation study by training the model on only four of the five datasets, and subsequently testing the model on the excluded one (as opposed to a standard random split in Figure~\ref{Figure 1}). As shown in Table \ref{tab:results}, GPT-4o fine-tuned (with training assertiveness scores rounded to one decimal point) originally achieves the lowest average MSE. Thus, GPT-4o appears well-suited to capturing the linguistic nuances that contribute to perceived assertiveness, even in transfer settings. We later found larger Llama models also achieve strong performance, even outperforming GPT-4o. Due to resource constraints, in subsequent sections we present our original results using GPT-4o as our primary tool. However, we note that the Llama models provide promising open-weight alternatives for measuring the assertiveness of generated explanations in the misinformation detection domain.

\begin{table}[ht]
\centering
\resizebox{0.9\linewidth}{!}{%
\begin{tabular}{lcccccc}
\toprule
\textbf{Model} & \textbf{Anthropic} & \textbf{Pei} & \textbf{LLama3-8b} & \textbf{GM} & \textbf{CMV} \\ 
\midrule
Base Pei & 1.91 & \textbf{0.83} & 1.56 & 1.92 & 2.31 \\ 
Fine-tuned Pei & 2.6 & 2.08 & 1.29 & 1.54 & 4.26 \\ 
Fine-tuned Llama-3.2-1B-Instruct & 1.85 & 2.14 & 2.05 & 2.06 & 1.79 \\ 
Fine-tuned Llama-3.2-3B-Instruct (Rounded) & 1.89 & 1.53 & 2.15 & 1.85 & 1.33 \\ 
Fine-tuned Llama-3.2-11B-Vision-Instruct (Rounded) & \textbf{0.90} & \textbf{0.83} & \textbf{0.92} & \textbf{0.82} & 0.90 \\ 
Prompted GPT & 1.07 & 1.42 & 1.90 & 1.16 & \textbf{0.75} \\ 
Fine-tuned GPT & 1.04 & 1.24 & 1.36 & 0.99 & 1.16 \\ 
Fine-tuned GPT (Rounded) & 0.99 & 1.05 & 1.42 & 0.98 & 0.94 \\ 
\bottomrule
\end{tabular}%
}
\caption{\small A comparison of model performance when training on four datasets and holding out the fifth for the measurements reported here. Numbers report Mean Squared Error (MSE) for each model. \\
\textbf{Initial Findings (November 2024)}: The fine-tuned GPT-4o model, using assertiveness scores rounded to one decimal place, achieves the best overall performance with an average MSE of approximately $1.078$. This result indicates that the model is the most consistent and reliable for assertiveness scoring across various domains. \\
\textbf{Followup Findings (March 2025)}: We add experiments with Fine-tuned Llama-3.2-3B and Llama-3.2-11B, with the same instructions as Fine-tuned GPT- models are added in this analysis with the same instruction as the previous Fine-tuned GPT-4o. Llama-3.2-11B achieves the best overall performance with an average MSE of approximately $0.87$. This provides a strong open-weight alternative to GPT-4o.}
\label{tab:results}
\end{table}

\begin{figure}[h]
\begin{subfigure}{\linewidth}
\centering
\includegraphics[width=0.7\linewidth]{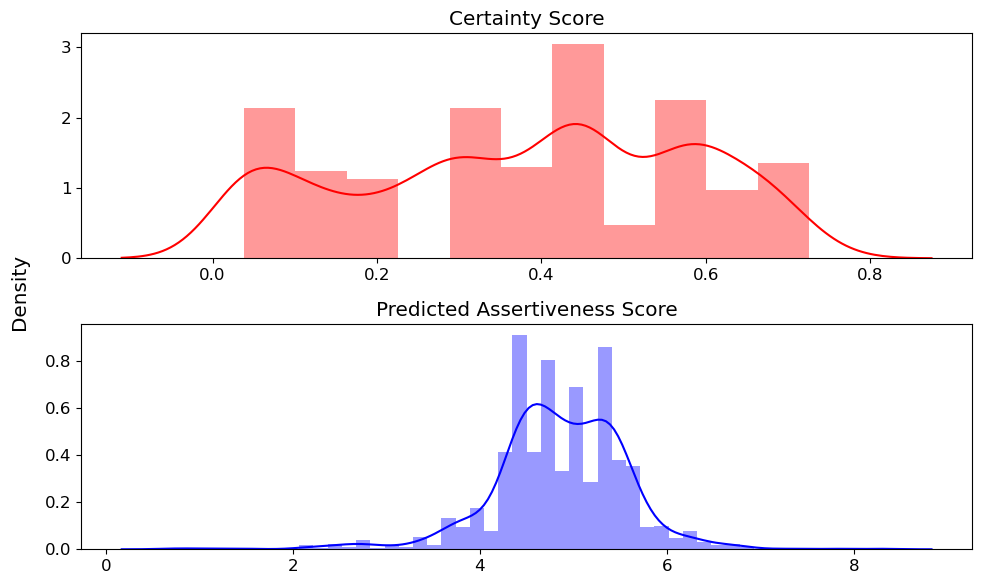}
    \caption{\small Certainty scores are relatively spread out; assertiveness scores are more concentrated towards the middle.}
    \label{fig:certainty-assertiveness-plots}
\end{subfigure}
\bigskip
\begin{subfigure}{\linewidth}
\centering
\includegraphics[width=0.7\linewidth]{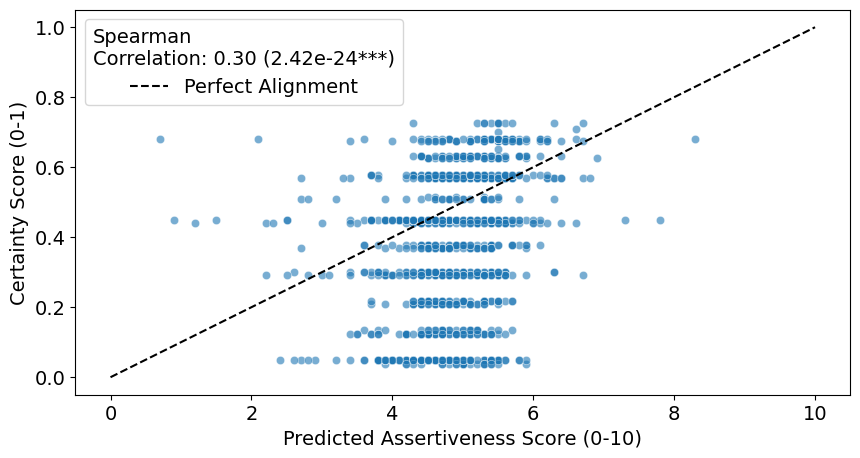}
    \caption{\small Low correlation (0.3) between the certainty score and model's predicted assertiveness score.}
    \label{fig:spearman_corr}
\end{subfigure}
\caption{Epistemic Miscalibration: misalignment between the LLM's certainty score and our model's assertiveness scores.}
\end{figure}

\paragraph{Certainty Calibration Score}
Figure \ref{fig:certainty-assertiveness-plots} illustrates a comparison between the probability distributions of certainty scores derived from the certainty calibration method proposed by \citet{pelrine2023reliable} and assertiveness scores from the best-performing model shown in Figure \ref{Figure 1} (Fine-tuned GPT, Rounded), applied to the LIAR misinformation dataset. Notably, while the certainty scores exhibit a wide variance, assertiveness scores are more concentrated toward the middle of the distribution. Additionally, Figure \ref{fig:spearman_corr} reveals a low Spearman correlation (0.3) between the two sets of scores, suggesting that there is a significant misalignment between certainty and assertiveness. We provide examples of both epistemically calibrated and uncalibrated explanations with varying levels of assertiveness in Appendix~\ref{Case Studies}. Furthermore, in Appendix~\ref{app:truefalseneither}, we test whether strong assertions of uncertainty affect calibration, finding that even when controlling for these cases, the model remains heavily skewed towards over-assertiveness.

\section{Human perceptions of linguistic assertiveness}
In the preceding sections, we provided empirical evidence highlighting the epistemic calibration problem. Our findings revealed a significant mismatch between the internal certainty and linguistic assertiveness of LLMs, especially in scenarios where their level of internal certainty is low (LLM exhibits high external assertiveness). 
However, for the epistemic calibration problem to be a strong normative issue, it is essential to establish that human (subjective) perceptions of linguistic assertiveness align with the assertiveness measurements obtained using our model. 
To address this critical validation step, we conducted an online survey to gather subjective assessments of assertiveness from 467 human respondents representative of a cross-section of the United States population. Participants were asked to evaluate the assertiveness of various explanations generated by GPT in a misinformation classification task.

\subsection{Description of the experiment} 
Respondents were presented with a series of statements, each accompanied by a true/false classification and an explanation generated by GPT. Participants were then instructed to rate the assertiveness of each explanation on a scale from 0 (Not at all assertive) to 10 (Extremely assertive). 
This task was repeated four times for each respondent, providing a dataset of 1868 human ratings of assertiveness. We provide more details including the prompt given to respondents in Appendix~\ref{app:survey}.

\paragraph{Explanation generation}
Initially, GPT-4 is prompted to provide a classification and explanation for each statement from the LIAR dataset, following the explain-then-score prompt in \citet{pelrine2023reliable} and other sections of this paper. We then prompt GPT-4o to generate two additional versions of each explanation: one less assertive and one more assertive than the original. This is to ensure that we have three distinct versions of explanations for each statement, allowing for meaningful comparisons of human perceptions.
Four randomly sampled explanations from this validation dataset are presented to each respondent for rating assertiveness.\footnote{We include in Appendix~\ref{app:survey} details about attention checks and prompting strategies used in our survey.}

\begin{table*}[h]
\caption{Spearman correlation between internal certainty, objective and subjective assertiveness}
\label{tab:correlations}
\begin{tabularx}{\linewidth}{lllll}
\toprule
 & Overall & Low & Medium & High \\
\midrule
Predicted Assertiveness vs. Human Assertiveness & 0.593*** & 0.048 & 0.400*** & 0.372*** \\
Internal Certainty vs. Predicted Assertiveness & 0.054 & -0.017 & 0.140 & 0.197 \\
Internal Certainty vs. Human Assertiveness & 0.145* & 0.080 & 0.194 & 0.219* \\
\bottomrule
\end{tabularx}
\end{table*}

\section{Results of human perceptions of assertiveness}

\begin{figure}[htp]
\begin{subfigure}{\linewidth}
\centering
    \includegraphics[width=0.6\linewidth]{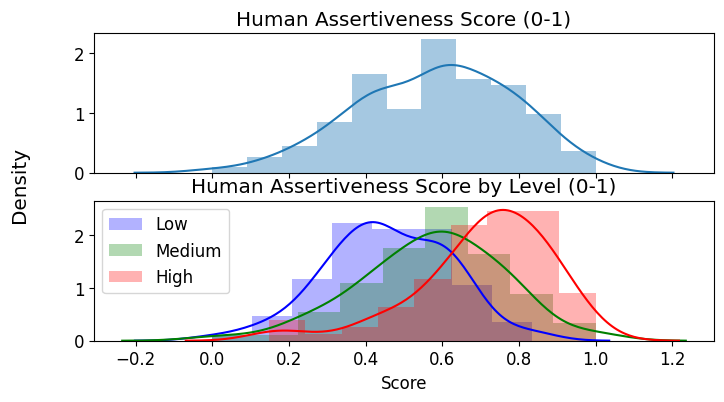}
    \caption{\small Mean assertiveness score distribution.}
    \label{Figure 4}
\end{subfigure}
\bigskip
\begin{subfigure}{\linewidth}
\centering
    \includegraphics[width=0.7\linewidth]{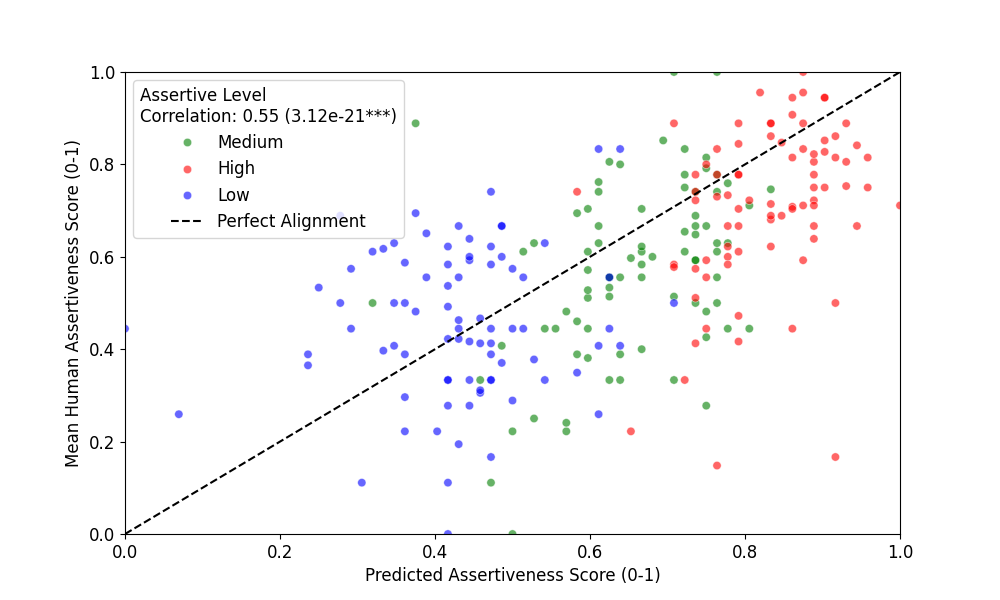}
    \caption{\small Correlation between predicted and human assertiveness scores.}
    \label{fig:corr_obj_sub}
\end{subfigure}
\bigskip
\begin{subfigure}{\linewidth}
\centering
    \includegraphics[width=0.8\linewidth]{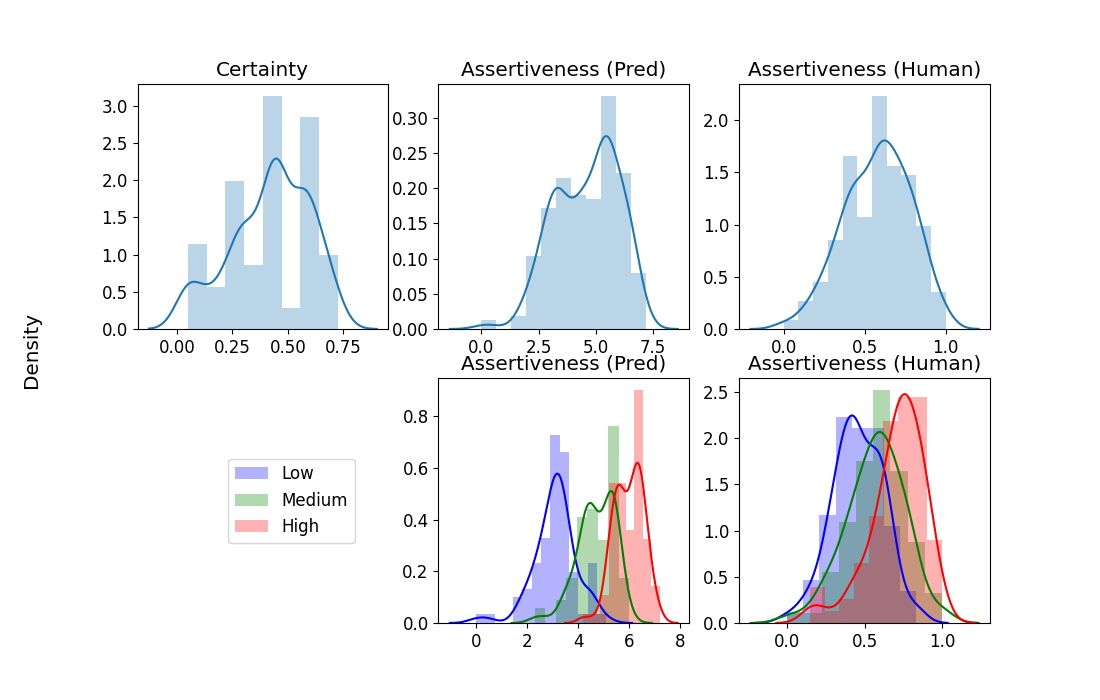}
    \caption{\small Internal certainty scores indicate the confidence level of the LLM in its own classification. Human assertiveness scores reflect how humans perceive the linguistic assertiveness of the explanations given by the LLM for its classifications.}
    \label{Figure 9}
\end{subfigure}

\caption{Our method's predicted assertiveness score is reasonably well aligned with the human scores, while the model's (internal) certainty is not.}

\end{figure}

In Figure \ref{Figure 4}, we observe that the scaled assertiveness scores from the survey are roughly normally distributed, centered at a score of 0.6. The scores  distributed across the three assertiveness levels (-1: low, 0: medium, 1: high), also shown in Figure~\ref{Figure 4}, confirms that our prompting strategy for generating explanations with different assertiveness levels is accurately perceived by human respondents. 


Figure \ref{fig:corr_obj_sub} plots assertiveness predicted by our model and the survey respondents, colored by assertiveness level, showing a strong relationship.

We also report the overall and disaggregated spearman correlations between both human and predicted assertiveness scores with internal certainty scores in Table \ref{tab:correlations} (correlations with p-values given in appendix Tables \ref{tab:app_pear_correlations} \& \ref{tab:app_spear_correlations}). The correlation between our model's predicted assertiveness scores and human perception of assertiveness is relatively strong at 0.55. This indicates that the predicted measures are fairly aligned with how humans perceive assertiveness. Meanwhile, the relationship between predicted assertiveness and internal certainty is very weak (0.064), highlighting the issue of epistemic miscalibration. Figure \ref{Figure 9} also shows comparisons between the internal certainty, predicted and human assertiveness scores.

\section{Discussion and Conclusion}
In this work, we discussed the problem of epistemic calibration for LLMs: ensuring that the confidence expressed in a model's communication aligns with its underlying reliability. We argued that this normative ideal is critical for LLMs to serve as robust and responsible information sources. Through a decomposition of the problem into external and internal certainty, we developed a framework for understanding and evaluating epistemic calibration in LLMs. Using our approach that greatly improves fidelity of assertiveness measurements compared to prior models, our empirical investigation of a state-of-the-art model reveals significant gaps between the model's internal confidence estimates and the assertiveness of its generated language. This miscalibration poses risks to users, who may be misled by overconfident model outputs. 

Our work also highlights the need for further research to fully understand and address the challenges of epistemic calibration. One key direction is developing new training and inference techniques to improve the alignment between LLMs' probability estimates and their linguistic expression of confidence. Another is studying the downstream impacts of epistemic miscalibration on user trust, decision making, and information ecosystems, through a combination of user studies and large-scale simulations. We believe that the epistemic calibration framework introduced in this paper provides a valuable foundation for these future efforts. 
We discuss further applications, to RLHF, silicon sampling, and debate, in Appendix~\ref{app:future}. 
Ultimately, achieving epistemic calibration in language models is not just a technical challenge, but a societal imperative. As these models become ever more integrated into our information-seeking and decision-making practices, ensuring that they express confidence in a calibrated and responsible way is essential for mitigating the risks of misinformation, confusion, and unwarranted trust. 

\section{Limitations}
Despite the promising findings and advancements discussed in this paper, several limitations should be acknowledged to provide a balanced perspective on the epistemic calibration of language models. First, our primary evaluation focuses on the directionality of variation in assertiveness and certainty. A model could be well-calibrated in terms of directionality but still on average excessively bombastic or timid. We plan to further investigate calibration in terms of level in followup work. Second, we do not experiment with the implications of epistemic miscalibration on the formation of human beliefs. Finally, while our study highlights the problem of epistemic calibration, it does not explore potential intervention strategies to mitigate this problem beyond the scope of our current
methods. We provide more details on the limitations of this study in section \ref{limitations} of the Appendix.


\section{Acknowledgements}
This work was partially funded by the CIFAR AI Chairs Program, the Centre for the Study of Democratic Citizenship, the Social Sciences and Humanities Research Council of Canada (grant no. 435-2023-0628), and the Digital Citizen Contribution Program of Canadian Heritage. We also thank Berkeley SPAR for connecting collaborators and funding support. Kellin Pelrine was supported by funding from IVADO and by the Fonds de recherche du Queb\'ec.

\section{Author Contributions}
Bijean Ghafouri led the human evaluation of assertivity, took the lead writing the second (current) version of the paper, and gave input and feedback throughout the project. Shahrad Mohammadzadeh led the work towards a stronger assertiveness measurement model, and made significant contributions to the direction, writing, and other parts of the project. James Zhou had the idea of assertiveness miscalibration, created the original framing and first version of the paper, and built and ran the initial experiments. Pratheeksha Nair conducted downstream analysis of assertiveness (mis)calibration and contributed to the writing. Jacob-Junqi Tian fine-tuned Llama-3.2-1B and set up the framework for our composite dataset labeling. Hikaru Tsujimura extended Llama fine-tuning experiments to 3B and 11B models. Mayank Goel expanded the assertivity evaluation experiments. Sukanya Krishna contributed to literature review and writing. Reihaneh Rabbany and Jean-Fran\c{c}ois Godbout advised the project, contributing ideas and feedback. Godbout also helped design the human evaluation and contributed substantially to the writing. Kellin Pelrine supervised the project, providing guidance and feedback at all stages.

\bibliography{main}

\begin{thebibliography}{53}
\providecommand{\natexlab}[1]{#1}
\providecommand{\url}[1]{\texttt{#1}}
\expandafter\ifx\csname urlstyle\endcsname\relax
  \providecommand{\doi}[1]{doi: #1}\else
  \providecommand{\doi}{doi: \begingroup \urlstyle{rm}\Url}\fi

\bibitem[Ali et~al.(2023)Ali, Dobbs, Hutchings, and Whitaker]{ali2023using}
Stephen~R Ali, Thomas~D Dobbs, Hayley~A Hutchings, and Iain~S Whitaker.
\newblock Using chatgpt to write patient clinic letters.
\newblock \emph{The Lancet Digital Health}, 5\penalty0 (4):\penalty0 e179--e181, 2023.

\bibitem[Argyle et~al.(2023)Argyle, Busby, Fulda, Gubler, Rytting, and Wingate]{Argyle_2023}
Lisa~P. Argyle, Ethan~C. Busby, Nancy Fulda, Joshua~R. Gubler, Christopher Rytting, and David Wingate.
\newblock Out of one, many: Using language models to simulate human samples.
\newblock \emph{Political Analysis}, 31\penalty0 (3):\penalty0 337–351, February 2023.
\newblock ISSN 1476-4989.
\newblock \doi{10.1017/pan.2023.2}.
\newblock URL \url{http://dx.doi.org/10.1017/pan.2023.2}.

\bibitem[Belem et~al.(2024)Belem, Kelly, Steyvers, Singh, and Smyth]{belem2024perceptions}
C.~G. Belem, M.~Kelly, M.~Steyvers, S.~Singh, and P.~Smyth.
\newblock Perceptions of linguistic uncertainty by language models and humans.
\newblock \emph{arXiv preprint arXiv:2407.15814}, 2024.

\bibitem[Breum et~al.(2023)Breum, Egdal, Mortensen, Møller, and Aiello]{breum2023persuasive}
Simon~Martin Breum, Daniel~Vædele Egdal, Victor~Gram Mortensen, Anders~Giovanni Møller, and Luca~Maria Aiello.
\newblock The persuasive power of large language models, 2023.

\bibitem[Budescu \& Wallsten(1985)Budescu and Wallsten]{budescu1985consistency}
David~V. Budescu and Thomas~S. Wallsten.
\newblock Consistency in interpretation of probabilistic phrases.
\newblock \emph{Organizational Behavior and Human Decision Processes}, 36\penalty0 (3):\penalty0 391--405, 1985.

\bibitem[Byalyk \& Nizhnik(2022)Byalyk and Nizhnik]{Byalyk_Nizhnik_2022}
Vasyl~Dmytrovych Byalyk and Liudmyla~Ivanivna Nizhnik.
\newblock Epistemic words on the confidence scale.
\newblock \emph{Academic Journal of Modern Philology}, \penalty0 (15):\penalty0 107–115, 2022.

\bibitem[Chan et~al.(2023)Chan, Chen, Su, Yu, Xue, Zhang, Fu, and Liu]{chan2023chateval}
Chi-Min Chan, Weize Chen, Yusheng Su, Jianxuan Yu, Wei Xue, Shanghang Zhang, Jie Fu, and Zhiyuan Liu.
\newblock Chateval: Towards better llm-based evaluators through multi-agent debate, 2023.

\bibitem[Chhikara(2025)]{chhikara2025mindconfidencegapoverconfidence}
Prateek Chhikara.
\newblock Mind the confidence gap: Overconfidence, calibration, and distractor effects in large language models, 2025.
\newblock URL \url{https://arxiv.org/abs/2502.11028}.

\bibitem[Clarke et~al.(1992)Clarke, Ruffin, Hill, and Beamen]{clarke1992ratings}
Valerie~A. Clarke, Coral~L. Ruffin, David~J. Hill, and Arthur~L. Beamen.
\newblock Ratings of orally presented verbal expressions of probability by a heterogeneous sample.
\newblock \emph{Journal of Applied Social Psychology}, 22\penalty0 (8):\penalty0 638--656, 1992.

\bibitem[Desai \& Durrett(2020)Desai and Durrett]{desai2020calibration}
Shrey Desai and Greg Durrett.
\newblock Calibration of pre-trained transformers.
\newblock \emph{arXiv preprint arXiv:2003.07892}, 2020.

\bibitem[Duan et~al.(2024)Duan, Cheng, Wang, Zavalny, Wang, Xu, Kailkhura, and Xu]{duan2024shifting}
Jinhao Duan, Hao Cheng, Shiqi Wang, Alex Zavalny, Chenan Wang, Renjing Xu, Bhavya Kailkhura, and Kaidi Xu.
\newblock Shifting attention to relevance: Towards the predictive uncertainty quantification of free-form large language models.
\newblock In \emph{Proceedings of the 62nd Annual Meeting of the Association for Computational Linguistics}, Bangkok, Thailand, Aug 2024. Association for Computational Linguistics.

\bibitem[Dubey et~al.(2024)Dubey, Jauhri, Pandey, Kadian, Al-Dahle, Letman, Mathur, Schelten, Yang, Fan, Goyal, Hartshorn, Yang, Mitra, Sravankumar, Korenev, Hinsvark, Rao, Zhang, Rodriguez, Gregerson, Spataru, Roziere, Biron, Tang, Chern, Caucheteux, Nayak, Bi, Marra, McConnell, Keller, Touret, Wu, Wong, Ferrer, Nikolaidis, Allonsius, Song, Pintz, Livshits, Esiobu, Choudhary, Mahajan, Garcia-Olano, Perino, Hupkes, Lakomkin, AlBadawy, Lobanova, Dinan, Smith, Radenovic, Zhang, Synnaeve, Lee, Anderson, Nail, Mialon, Pang, Cucurell, Nguyen, Korevaar, Xu, Touvron, Zarov, Ibarra, Kloumann, Misra, Evtimov, Copet, Lee, Geffert, Vranes, Park, Mahadeokar, Shah, van~der Linde, Billock, Hong, Lee, Fu, Chi, Huang, Liu, Wang, Yu, Bitton, Spisak, Park, Rocca, Johnstun, Saxe, Jia, Alwala, Upasani, Plawiak, Li, Heafield, Stone, El-Arini, Iyer, Malik, Chiu, Bhalla, Rantala-Yeary, van~der Maaten, Chen, Tan, Jenkins, Martin, Madaan, Malo, Blecher, Landzaat, de~Oliveira, Muzzi, Pasupuleti, Singh, Paluri, Kardas, Oldham, Rita,
  Pavlova, Kambadur, Lewis, Si, Singh, Hassan, Goyal, Torabi, Bashlykov, Bogoychev, Chatterji, Duchenne, Çelebi, Alrassy, Zhang, Li, Vasic, Weng, Bhargava, Dubal, Krishnan, Koura, Xu, He, Dong, Srinivasan, Ganapathy, Calderer, Cabral, Stojnic, Raileanu, Girdhar, Patel, Sauvestre, Polidoro, Sumbaly, Taylor, Silva, Hou, Wang, Hosseini, Chennabasappa, Singh, Bell, Kim, Edunov, Nie, Narang, Raparthy, Shen, Wan, Bhosale, Zhang, Vandenhende, Batra, Whitman, Sootla, Collot, Gururangan, Borodinsky, Herman, Fowler, Sheasha, Georgiou, Scialom, Speckbacher, Mihaylov, Xiao, Karn, Goswami, Gupta, Ramanathan, Kerkez, Gonguet, Do, Vogeti, Petrovic, Chu, Xiong, Fu, Meers, Martinet, Wang, Tan, Xie, Jia, Wang, Goldschlag, Gaur, Babaei, Wen, Song, Zhang, Li, Mao, Coudert, Yan, Chen, Papakipos, Singh, Grattafiori, Jain, Kelsey, Shajnfeld, Gangidi, Victoria, Goldstand, Menon, Sharma, Boesenberg, Vaughan, Baevski, Feinstein, Kallet, Sangani, Yunus, Lupu, Alvarado, Caples, Gu, Ho, Poulton, Ryan, Ramchandani, Franco, Saraf,
  Chowdhury, Gabriel, Bharambe, Eisenman, Yazdan, James, Maurer, Leonhardi, Huang, Loyd, Paola, Paranjape, Liu, Wu, Ni, Hancock, Wasti, Spence, Stojkovic, Gamido, Montalvo, Parker, Burton, Mejia, Wang, Kim, Zhou, Hu, Chu, Cai, Tindal, Feichtenhofer, Civin, Beaty, Kreymer, Li, Wyatt, Adkins, Xu, Testuggine, David, Parikh, Liskovich, Foss, Wang, Le, Holland, Dowling, Jamil, Montgomery, Presani, Hahn, Wood, Brinkman, Arcaute, Dunbar, Smothers, Sun, Kreuk, Tian, Ozgenel, Caggioni, Guzmán, Kanayet, Seide, Florez, Schwarz, Badeer, Swee, Halpern, Thattai, Herman, Sizov, Guangyi, Zhang, Lakshminarayanan, Shojanazeri, Zou, Wang, Zha, Habeeb, Rudolph, Suk, Aspegren, Goldman, Damlaj, Molybog, Tufanov, Veliche, Gat, Weissman, Geboski, Kohli, Asher, Gaya, Marcus, Tang, Chan, Zhen, Reizenstein, Teboul, Zhong, Jin, Yang, Cummings, Carvill, Shepard, McPhie, Torres, Ginsburg, Wang, Wu, U, Saxena, Prasad, Khandelwal, Zand, Matosich, Veeraraghavan, Michelena, Li, Huang, Chawla, Lakhotia, Huang, Chen, Garg, A, Silva, Bell,
  Zhang, Guo, Yu, Moshkovich, Wehrstedt, Khabsa, Avalani, Bhatt, Tsimpoukelli, Mankus, Hasson, Lennie, Reso, Groshev, Naumov, Lathi, Keneally, Seltzer, Valko, Restrepo, Patel, Vyatskov, Samvelyan, Clark, Macey, Wang, Hermoso, Metanat, Rastegari, Bansal, Santhanam, Parks, White, Bawa, Singhal, Egebo, Usunier, Laptev, Dong, Zhang, Cheng, Chernoguz, Hart, Salpekar, Kalinli, Kent, Parekh, Saab, Balaji, Rittner, Bontrager, Roux, Dollar, Zvyagina, Ratanchandani, Yuvraj, Liang, Alao, Rodriguez, Ayub, Murthy, Nayani, Mitra, Li, Hogan, Battey, Wang, Maheswari, Howes, Rinott, Bondu, Datta, Chugh, Hunt, Dhillon, Sidorov, Pan, Verma, Yamamoto, Ramaswamy, Lindsay, Lindsay, Feng, Lin, Zha, Shankar, Zhang, Zhang, Wang, Agarwal, Sajuyigbe, Chintala, Max, Chen, Kehoe, Satterfield, Govindaprasad, Gupta, Cho, Virk, Subramanian, Choudhury, Goldman, Remez, Glaser, Best, Kohler, Robinson, Li, Zhang, Matthews, Chou, Shaked, Vontimitta, Ajayi, Montanez, Mohan, Kumar, Mangla, Albiero, Ionescu, Poenaru, Mihailescu, Ivanov, Li, Wang,
  Jiang, Bouaziz, Constable, Tang, Wang, Wu, Wang, Xia, Wu, Gao, Chen, Hu, Jia, Qi, Li, Zhang, Zhang, Adi, Nam, Yu, Wang, Hao, Qian, He, Rait, DeVito, Rosnbrick, Wen, Yang, and Zhao]{dubey2024llama3herdmodels}
Abhimanyu Dubey, Abhinav Jauhri, Abhinav Pandey, Abhishek Kadian, Ahmad Al-Dahle, Aiesha Letman, Akhil Mathur, Alan Schelten, Amy Yang, Angela Fan, Anirudh Goyal, Anthony Hartshorn, Aobo Yang, Archi Mitra, Archie Sravankumar, Artem Korenev, Arthur Hinsvark, Arun Rao, Aston Zhang, Aurelien Rodriguez, Austen Gregerson, Ava Spataru, Baptiste Roziere, Bethany Biron, Binh Tang, Bobbie Chern, Charlotte Caucheteux, Chaya Nayak, Chloe Bi, Chris Marra, Chris McConnell, Christian Keller, Christophe Touret, Chunyang Wu, Corinne Wong, Cristian~Canton Ferrer, Cyrus Nikolaidis, Damien Allonsius, Daniel Song, Danielle Pintz, Danny Livshits, David Esiobu, Dhruv Choudhary, Dhruv Mahajan, Diego Garcia-Olano, Diego Perino, Dieuwke Hupkes, Egor Lakomkin, Ehab AlBadawy, Elina Lobanova, Emily Dinan, Eric~Michael Smith, Filip Radenovic, Frank Zhang, Gabriel Synnaeve, Gabrielle Lee, Georgia~Lewis Anderson, Graeme Nail, Gregoire Mialon, Guan Pang, Guillem Cucurell, Hailey Nguyen, Hannah Korevaar, Hu~Xu, Hugo Touvron, Iliyan Zarov,
  Imanol~Arrieta Ibarra, Isabel Kloumann, Ishan Misra, Ivan Evtimov, Jade Copet, Jaewon Lee, Jan Geffert, Jana Vranes, Jason Park, Jay Mahadeokar, Jeet Shah, Jelmer van~der Linde, Jennifer Billock, Jenny Hong, Jenya Lee, Jeremy Fu, Jianfeng Chi, Jianyu Huang, Jiawen Liu, Jie Wang, Jiecao Yu, Joanna Bitton, Joe Spisak, Jongsoo Park, Joseph Rocca, Joshua Johnstun, Joshua Saxe, Junteng Jia, Kalyan~Vasuden Alwala, Kartikeya Upasani, Kate Plawiak, Ke~Li, Kenneth Heafield, Kevin Stone, Khalid El-Arini, Krithika Iyer, Kshitiz Malik, Kuenley Chiu, Kunal Bhalla, Lauren Rantala-Yeary, Laurens van~der Maaten, Lawrence Chen, Liang Tan, Liz Jenkins, Louis Martin, Lovish Madaan, Lubo Malo, Lukas Blecher, Lukas Landzaat, Luke de~Oliveira, Madeline Muzzi, Mahesh Pasupuleti, Mannat Singh, Manohar Paluri, Marcin Kardas, Mathew Oldham, Mathieu Rita, Maya Pavlova, Melanie Kambadur, Mike Lewis, Min Si, Mitesh~Kumar Singh, Mona Hassan, Naman Goyal, Narjes Torabi, Nikolay Bashlykov, Nikolay Bogoychev, Niladri Chatterji, Olivier
  Duchenne, Onur Çelebi, Patrick Alrassy, Pengchuan Zhang, Pengwei Li, Petar Vasic, Peter Weng, Prajjwal Bhargava, Pratik Dubal, Praveen Krishnan, Punit~Singh Koura, Puxin Xu, Qing He, Qingxiao Dong, Ragavan Srinivasan, Raj Ganapathy, Ramon Calderer, Ricardo~Silveira Cabral, Robert Stojnic, Roberta Raileanu, Rohit Girdhar, Rohit Patel, Romain Sauvestre, Ronnie Polidoro, Roshan Sumbaly, Ross Taylor, Ruan Silva, Rui Hou, Rui Wang, Saghar Hosseini, Sahana Chennabasappa, Sanjay Singh, Sean Bell, Seohyun~Sonia Kim, Sergey Edunov, Shaoliang Nie, Sharan Narang, Sharath Raparthy, Sheng Shen, Shengye Wan, Shruti Bhosale, Shun Zhang, Simon Vandenhende, Soumya Batra, Spencer Whitman, Sten Sootla, Stephane Collot, Suchin Gururangan, Sydney Borodinsky, Tamar Herman, Tara Fowler, Tarek Sheasha, Thomas Georgiou, Thomas Scialom, Tobias Speckbacher, Todor Mihaylov, Tong Xiao, Ujjwal Karn, Vedanuj Goswami, Vibhor Gupta, Vignesh Ramanathan, Viktor Kerkez, Vincent Gonguet, Virginie Do, Vish Vogeti, Vladan Petrovic, Weiwei Chu,
  Wenhan Xiong, Wenyin Fu, Whitney Meers, Xavier Martinet, Xiaodong Wang, Xiaoqing~Ellen Tan, Xinfeng Xie, Xuchao Jia, Xuewei Wang, Yaelle Goldschlag, Yashesh Gaur, Yasmine Babaei, Yi~Wen, Yiwen Song, Yuchen Zhang, Yue Li, Yuning Mao, Zacharie~Delpierre Coudert, Zheng Yan, Zhengxing Chen, Zoe Papakipos, Aaditya Singh, Aaron Grattafiori, Abha Jain, Adam Kelsey, Adam Shajnfeld, Adithya Gangidi, Adolfo Victoria, Ahuva Goldstand, Ajay Menon, Ajay Sharma, Alex Boesenberg, Alex Vaughan, Alexei Baevski, Allie Feinstein, Amanda Kallet, Amit Sangani, Anam Yunus, Andrei Lupu, Andres Alvarado, Andrew Caples, Andrew Gu, Andrew Ho, Andrew Poulton, Andrew Ryan, Ankit Ramchandani, Annie Franco, Aparajita Saraf, Arkabandhu Chowdhury, Ashley Gabriel, Ashwin Bharambe, Assaf Eisenman, Azadeh Yazdan, Beau James, Ben Maurer, Benjamin Leonhardi, Bernie Huang, Beth Loyd, Beto~De Paola, Bhargavi Paranjape, Bing Liu, Bo~Wu, Boyu Ni, Braden Hancock, Bram Wasti, Brandon Spence, Brani Stojkovic, Brian Gamido, Britt Montalvo, Carl
  Parker, Carly Burton, Catalina Mejia, Changhan Wang, Changkyu Kim, Chao Zhou, Chester Hu, Ching-Hsiang Chu, Chris Cai, Chris Tindal, Christoph Feichtenhofer, Damon Civin, Dana Beaty, Daniel Kreymer, Daniel Li, Danny Wyatt, David Adkins, David Xu, Davide Testuggine, Delia David, Devi Parikh, Diana Liskovich, Didem Foss, Dingkang Wang, Duc Le, Dustin Holland, Edward Dowling, Eissa Jamil, Elaine Montgomery, Eleonora Presani, Emily Hahn, Emily Wood, Erik Brinkman, Esteban Arcaute, Evan Dunbar, Evan Smothers, Fei Sun, Felix Kreuk, Feng Tian, Firat Ozgenel, Francesco Caggioni, Francisco Guzmán, Frank Kanayet, Frank Seide, Gabriela~Medina Florez, Gabriella Schwarz, Gada Badeer, Georgia Swee, Gil Halpern, Govind Thattai, Grant Herman, Grigory Sizov, Guangyi, Zhang, Guna Lakshminarayanan, Hamid Shojanazeri, Han Zou, Hannah Wang, Hanwen Zha, Haroun Habeeb, Harrison Rudolph, Helen Suk, Henry Aspegren, Hunter Goldman, Ibrahim Damlaj, Igor Molybog, Igor Tufanov, Irina-Elena Veliche, Itai Gat, Jake Weissman, James
  Geboski, James Kohli, Japhet Asher, Jean-Baptiste Gaya, Jeff Marcus, Jeff Tang, Jennifer Chan, Jenny Zhen, Jeremy Reizenstein, Jeremy Teboul, Jessica Zhong, Jian Jin, Jingyi Yang, Joe Cummings, Jon Carvill, Jon Shepard, Jonathan McPhie, Jonathan Torres, Josh Ginsburg, Junjie Wang, Kai Wu, Kam~Hou U, Karan Saxena, Karthik Prasad, Kartikay Khandelwal, Katayoun Zand, Kathy Matosich, Kaushik Veeraraghavan, Kelly Michelena, Keqian Li, Kun Huang, Kunal Chawla, Kushal Lakhotia, Kyle Huang, Lailin Chen, Lakshya Garg, Lavender A, Leandro Silva, Lee Bell, Lei Zhang, Liangpeng Guo, Licheng Yu, Liron Moshkovich, Luca Wehrstedt, Madian Khabsa, Manav Avalani, Manish Bhatt, Maria Tsimpoukelli, Martynas Mankus, Matan Hasson, Matthew Lennie, Matthias Reso, Maxim Groshev, Maxim Naumov, Maya Lathi, Meghan Keneally, Michael~L. Seltzer, Michal Valko, Michelle Restrepo, Mihir Patel, Mik Vyatskov, Mikayel Samvelyan, Mike Clark, Mike Macey, Mike Wang, Miquel~Jubert Hermoso, Mo~Metanat, Mohammad Rastegari, Munish Bansal, Nandhini
  Santhanam, Natascha Parks, Natasha White, Navyata Bawa, Nayan Singhal, Nick Egebo, Nicolas Usunier, Nikolay~Pavlovich Laptev, Ning Dong, Ning Zhang, Norman Cheng, Oleg Chernoguz, Olivia Hart, Omkar Salpekar, Ozlem Kalinli, Parkin Kent, Parth Parekh, Paul Saab, Pavan Balaji, Pedro Rittner, Philip Bontrager, Pierre Roux, Piotr Dollar, Polina Zvyagina, Prashant Ratanchandani, Pritish Yuvraj, Qian Liang, Rachad Alao, Rachel Rodriguez, Rafi Ayub, Raghotham Murthy, Raghu Nayani, Rahul Mitra, Raymond Li, Rebekkah Hogan, Robin Battey, Rocky Wang, Rohan Maheswari, Russ Howes, Ruty Rinott, Sai~Jayesh Bondu, Samyak Datta, Sara Chugh, Sara Hunt, Sargun Dhillon, Sasha Sidorov, Satadru Pan, Saurabh Verma, Seiji Yamamoto, Sharadh Ramaswamy, Shaun Lindsay, Shaun Lindsay, Sheng Feng, Shenghao Lin, Shengxin~Cindy Zha, Shiva Shankar, Shuqiang Zhang, Shuqiang Zhang, Sinong Wang, Sneha Agarwal, Soji Sajuyigbe, Soumith Chintala, Stephanie Max, Stephen Chen, Steve Kehoe, Steve Satterfield, Sudarshan Govindaprasad, Sumit Gupta,
  Sungmin Cho, Sunny Virk, Suraj Subramanian, Sy~Choudhury, Sydney Goldman, Tal Remez, Tamar Glaser, Tamara Best, Thilo Kohler, Thomas Robinson, Tianhe Li, Tianjun Zhang, Tim Matthews, Timothy Chou, Tzook Shaked, Varun Vontimitta, Victoria Ajayi, Victoria Montanez, Vijai Mohan, Vinay~Satish Kumar, Vishal Mangla, Vítor Albiero, Vlad Ionescu, Vlad Poenaru, Vlad~Tiberiu Mihailescu, Vladimir Ivanov, Wei Li, Wenchen Wang, Wenwen Jiang, Wes Bouaziz, Will Constable, Xiaocheng Tang, Xiaofang Wang, Xiaojian Wu, Xiaolan Wang, Xide Xia, Xilun Wu, Xinbo Gao, Yanjun Chen, Ye~Hu, Ye~Jia, Ye~Qi, Yenda Li, Yilin Zhang, Ying Zhang, Yossi Adi, Youngjin Nam, Yu, Wang, Yuchen Hao, Yundi Qian, Yuzi He, Zach Rait, Zachary DeVito, Zef Rosnbrick, Zhaoduo Wen, Zhenyu Yang, and Zhiwei Zhao.
\newblock The llama 3 herd of models, 2024.
\newblock URL \url{https://arxiv.org/abs/2407.21783}.

\bibitem[Durmus et~al.(2024)Durmus, Lovitt, Tamkin, Ritchie, Clark, and Ganguli]{durmus2024persuasion}
Esin Durmus, Liane Lovitt, Alex Tamkin, Stuart Ritchie, Jack Clark, and Deep Ganguli.
\newblock Measuring the persuasiveness of language models, 2024.
\newblock URL \url{https://www.anthropic.com/news/measuring-model-persuasiveness}.
\newblock Accessed: 2024-04-09.

\bibitem[Erev \& Cohen(1990)Erev and Cohen]{erev1990verbal}
Ido Erev and Brent~L Cohen.
\newblock Verbal versus numerical probabilities: Efficiency, biases, and the preference paradox.
\newblock \emph{Organizational Behavior and Human Decision Processes}, 45\penalty0 (1):\penalty0 1--18, February 1990.
\newblock \doi{10.1016/0749-5978(90)90002-q}.
\newblock URL \url{http://dx.doi.org/10.1016/0749-5978(90)90002-Q}.

\bibitem[Hendrycks et~al.(2021)Hendrycks, Burns, Basart, Zou, Mazeika, Song, and Steinhardt]{hendrycks2021massive}
Dan Hendrycks, Collin Burns, Steven Basart, Andy Zou, Mantas Mazeika, Dawn Song, and Jacob Steinhardt.
\newblock Measuring massive multitask language understanding.
\newblock In \emph{Proceedings of the International Conference on Learning Representations (ICLR)}, 2021.

\bibitem[Hosking et~al.(2024)Hosking, Blunsom, and Bartolo]{hosking2024human}
Tom Hosking, Phil Blunsom, and Max Bartolo.
\newblock Human feedback is not gold standard, 2024.

\bibitem[Hu et~al.(2021)Hu, Shen, Wallis, Allen-Zhu, Li, Wang, Wang, and Chen]{hu2021loralowrankadaptationlarge}
Edward~J. Hu, Yelong Shen, Phillip Wallis, Zeyuan Allen-Zhu, Yuanzhi Li, Shean Wang, Lu~Wang, and Weizhu Chen.
\newblock Lora: Low-rank adaptation of large language models, 2021.
\newblock URL \url{https://arxiv.org/abs/2106.09685}.

\bibitem[Huang et~al.(2023)Huang, Yu, Ma, Zhong, Feng, Wang, Chen, Peng, Feng, and et~al.]{huang2023survey}
Lei Huang, Weijiang Yu, Weitao Ma, Weihong Zhong, Zhangyin Feng, Haotian Wang, Qianglong Chen, Weihua Peng, Xiaocheng Feng, and Bing~Qin et~al.
\newblock A survey on hallucination in large language models: Principles, taxonomy, challenges, and open questions, 2023.

\bibitem[Jiang et~al.(2021)Jiang, Araki, Ding, and Neubig]{jiang2021how}
Zhengbao Jiang, Jun Araki, Haibo Ding, and Graham Neubig.
\newblock How can we know when language models know? on the calibration of language models for question answering.
\newblock \emph{Transactions of the Association for Computational Linguistics}, 9:\penalty0 962--977, 2021.
\newblock \doi{10.1162/tacl_a_00407}.
\newblock URL \url{https://aclanthology.org/2021.tacl-1.57}.

\bibitem[Jo(2023)]{jo2023promise}
A~Jo.
\newblock The promise and peril of generative ai.
\newblock \emph{Nature}, 614\penalty0 (1):\penalty0 214--216, 2023.

\bibitem[Karelitz \& Budescu(2004)Karelitz and Budescu]{karelitz2004you}
Tzur~M. Karelitz and David~V. Budescu.
\newblock You say "probable" and i say "likely": Improving interpersonal communication with verbal probability phrases.
\newblock \emph{Journal of Experimental Psychology: Applied}, 10\penalty0 (1):\penalty0 25--41, 2004.
\newblock ISSN 1076-898X.
\newblock \doi{10.1037/1076-898X.10.1.25}.
\newblock URL \url{http://dx.doi.org/10.1037/1076-898X.10.1.25}.

\bibitem[Kim et~al.(2024)Kim, Lee, Huang, Chan, Li, and Ji]{kim2024llms}
Kyungha Kim, Sangyun Lee, Kung-Hsiang Huang, Hou~Pong Chan, Manling Li, and Heng Ji.
\newblock Can llms produce faithful explanations for fact-checking? towards faithful explainable fact-checking via multi-agent debate, 2024.

\bibitem[Kolhatkar et~al.(2020)Kolhatkar, Wu, Cavasso, Francis, Shukla, and Taboada]{kolhatkar2020sfu}
Varada Kolhatkar, Hanhan Wu, Luca Cavasso, Emilie Francis, Kavan Shukla, and Maite Taboada.
\newblock The sfu opinion and comments corpus: A corpus for the analysis of online news comments.
\newblock \emph{Corpus pragmatics}, 4:\penalty0 155--190, 2020.

\bibitem[Kuhn et~al.(2023)Kuhn, Gal, and Farquhar]{kuhn2023semantic}
Lorenz Kuhn, Yarin Gal, and Sebastian Farquhar.
\newblock Semantic uncertainty: Linguistic invariances for uncertainty estimation in natural language generation.
\newblock In \emph{Proceedings of the 11th International Conference on Learning Representations, ICLR'23}, 2023.

\bibitem[Lee et~al.(2023)Lee, Phatale, Mansoor, Mesnard, Ferret, Lu, Bishop, Hall, Carbune, Rastogi, and Prakash]{lee2023rlaif}
Harrison Lee, Samrat Phatale, Hassan Mansoor, Thomas Mesnard, Johan Ferret, Kellie Lu, Colton Bishop, Ethan Hall, Victor Carbune, Abhinav Rastogi, and Sushant Prakash.
\newblock Rlaif: Scaling reinforcement learning from human feedback with ai feedback, 2023.

\bibitem[Leng et~al.(2025)Leng, Huang, Zhu, and Huang]{leng2025tamingoverconfidencellmsreward}
Jixuan Leng, Chengsong Huang, Banghua Zhu, and Jiaxin Huang.
\newblock Taming overconfidence in llms: Reward calibration in rlhf, 2025.
\newblock URL \url{https://arxiv.org/abs/2410.09724}.

\bibitem[Lin et~al.(2022)Lin, Hilton, and Evans]{lin2022teaching}
Stephanie Lin, Jacob Hilton, and Owain Evans.
\newblock Teaching models to express their uncertainty in words.
\newblock \emph{Transactions on Machine Learning Research}, 2022.
\newblock ISSN 2835-8856.
\newblock URL \url{https://openreview.net/forum?id=8s8K2UZGTZ}.

\bibitem[Mielke et~al.(2022)Mielke, Szlam, Dinan, and Boureau]{mielke2022reducing}
Stephanie~J. Mielke, Alexander Szlam, Emily Dinan, and Yann~LeCun Boureau.
\newblock Reducing conversational agents’ overconfidence through linguistic calibration.
\newblock \emph{Transactions of the Association for Computational Linguistics}, 10:\penalty0 857--872, 2022.

\bibitem[OpenAI(2020)]{hello_GPT4o}
OpenAI.
\newblock Hello {GPT}-4o, 2020.
\newblock URL \url{https://openai.com/index/hello-gpt-4o/}.

\bibitem[Pei \& Jurgens(2021)Pei and Jurgens]{pei2021measuring}
Jiaxin Pei and David Jurgens.
\newblock Measuring sentence-level and aspect-level (un)certainty in science communications, 2021.

\bibitem[Pelrine et~al.(2023)Pelrine, Imouza, Thibault, Reksoprodjo, Gupta, Christoph, Godbout, and Rabbany]{pelrine2023reliable}
Kellin Pelrine, Anne Imouza, Camille Thibault, Meilina Reksoprodjo, Caleb Gupta, Joel Christoph, Jean-François Godbout, and Reihaneh Rabbany.
\newblock Towards reliable misinformation mitigation: Generalization, uncertainty, and gpt-4, 2023.

\bibitem[Platt et~al.(1999)]{platt1999probabilistic}
John Platt et~al.
\newblock Probabilistic outputs for support vector machines and comparisons to regularized likelihood methods.
\newblock \emph{Advances in large margin classifiers}, 10\penalty0 (3):\penalty0 61--74, 1999.

\bibitem[Ren et~al.(2023)Ren, Wang, Qu, Zhao, Liu, Tian, Wu, Wen, and Wang]{ren2023investigating}
Ruiyang Ren, Yuhao Wang, Yingqi Qu, Wayne~Xin Zhao, Jing Liu, Hao Tian, Hua Wu, Ji-Rong Wen, and Haifeng Wang.
\newblock Investigating the factual knowledge boundary of large language models with retrieval augmentation.
\newblock \emph{arXiv preprint arXiv:2307.11019}, 2023.

\bibitem[Rivera et~al.(2024)Rivera, Godbout, Rabbany, and Pelrine]{rivera2024combining}
Mauricio Rivera, Jean-Fran{\c{c}}ois Godbout, Reihaneh Rabbany, and Kellin Pelrine.
\newblock Combining confidence elicitation and sample-based methods for uncertainty quantification in misinformation mitigation.
\newblock \emph{arXiv preprint arXiv:2401.08694}, 2024.

\bibitem[Shrivastava et~al.(2023)Shrivastava, Liang, and Kumar]{shrivastava2023llamas}
Vaishnavi Shrivastava, Percy Liang, and Ananya Kumar.
\newblock Llamas know what gpts don’t show: Surrogate models for confidence estimation.
\newblock \emph{ArXiv}, abs/2311.08877, 2023.
\newblock URL \url{https://api.semanticscholar.org/CorpusID:265213392}.

\bibitem[Si et~al.(2022)Si, Gan, Yang, Wang, Wang, Boyd-Graber, and Wang]{si2022prompting}
Chenglei Si, Zhe Gan, Zhengyuan Yang, Shuohang Wang, Jianfeng Wang, Jordan Boyd-Graber, and Lijuan Wang.
\newblock Prompting gpt-3 to be reliable.
\newblock \emph{arXiv preprint arXiv:2210.09150}, 2022.

\bibitem[Sileo \& Moens(2023)Sileo and Moens]{sileo2023probing}
Damien Sileo and Marie-Francine Moens.
\newblock Probing neural language models for understanding of words of estimative probability.
\newblock In Alexis Palmer and Jose Camacho-Collados (eds.), \emph{Proceedings of the 12th Joint Conference on Lexical and Computational Semantics (*SEM 2023)}, pp.\  469--476, Toronto, Canada, July 2023. Association for Computational Linguistics.
\newblock \doi{10.18653/v1/2023.starsem-1.41}.
\newblock URL \url{https://aclanthology.org/2023.starsem-1.41}.

\bibitem[Steyvers et~al.(2024)Steyvers, Tejeda, Kumar, Belem, Karny, Hu, Mayer, and Smyth]{steyvers2024calibration}
Mark Steyvers, Heliodoro Tejeda, Aakriti Kumar, Catarina Belem, Sheer Karny, Xinyue Hu, Lukas Mayer, and Padhraic Smyth.
\newblock The calibration gap between model and human confidence in large language models.
\newblock \emph{arXiv preprint arXiv:2401.13835}, 2024.

\bibitem[von Werra et~al.(2020)von Werra, Belkada, Tunstall, Beeching, Thrush, Lambert, Huang, Rasul, and Gallouédec]{vonwerra2022trl}
Leandro von Werra, Younes Belkada, Lewis Tunstall, Edward Beeching, Tristan Thrush, Nathan Lambert, Shengyi Huang, Kashif Rasul, and Quentin Gallouédec.
\newblock Trl: Transformer reinforcement learning.
\newblock \url{https://github.com/huggingface/trl}, 2020.

\bibitem[Wallsten et~al.(1993)Wallsten, Budescu, Zwick, and Kemp]{wallsten1993preferences}
Thomas~S. Wallsten, David~V. Budescu, Rami Zwick, and Steven~M. Kemp.
\newblock Preferences and reasons for communicating probabilistic information in verbal or numerical terms.
\newblock \emph{Bulletin of the Psychonomic Society}, 31\penalty0 (2):\penalty0 135--138, February 1993.
\newblock \doi{10.3758/bf03334162}.
\newblock URL \url{http://dx.doi.org/10.3758/BF03334162}.

\bibitem[Wang(2017)]{wang2017liar}
William~Yang Wang.
\newblock "liar, liar pants on fire": A new benchmark dataset for fake news detection, 2017.

\bibitem[Wen et~al.(2024)Wen, Xu, HAN, Wolfe, Wang, and Howe]{wen2024from}
Bingbing Wen, Chenjun Xu, Bin HAN, Robert Wolfe, Lucy~Lu Wang, and Bill Howe.
\newblock From human to model overconfidence: Evaluating confidence dynamics in large language models.
\newblock In \emph{NeurIPS 2024 Workshop on Behavioral Machine Learning}, 2024.
\newblock URL \url{https://openreview.net/forum?id=y9UdO5cmHs}.

\bibitem[Whalen \& et~al.(2023)Whalen and et~al.]{whalen2023chatgpt}
Jeromie Whalen and Chrystalla~Mouza et~al.
\newblock Chatgpt: Challenges, opportunities, and implications for teacher education.
\newblock \emph{Contemporary Issues in Technology and Teacher Education}, 23\penalty0 (1):\penalty0 1--23, 2023.

\bibitem[Wiegmann et~al.(2022)Wiegmann, Al~Khatib, Khanna, and Stein]{wiegmann-etal-2022-analyzing}
Matti Wiegmann, Khalid Al~Khatib, Vishal Khanna, and Benno Stein.
\newblock Analyzing persuasion strategies of debaters on social media.
\newblock In Nicoletta Calzolari, Chu-Ren Huang, Hansaem Kim, James Pustejovsky, Leo Wanner, Key-Sun Choi, Pum-Mo Ryu, Hsin-Hsi Chen, Lucia Donatelli, Heng Ji, Sadao Kurohashi, Patrizia Paggio, Nianwen Xue, Seokhwan Kim, Younggyun Hahm, Zhong He, Tony~Kyungil Lee, Enrico Santus, Francis Bond, and Seung-Hoon Na (eds.), \emph{Proceedings of the 29th International Conference on Computational Linguistics}, pp.\  6897--6905, Gyeongju, Republic of Korea, October 2022. International Committee on Computational Linguistics.
\newblock URL \url{https://aclanthology.org/2022.coling-1.600}.

\bibitem[Windschitl \& Wells(1996)Windschitl and Wells]{windschitl1996measuring}
Paul~D. Windschitl and Gary~L. Wells.
\newblock Measuring psychological uncertainty: Verbal versus numeric methods.
\newblock \emph{Journal of Experimental Psychology: Applied}, 2\penalty0 (4):\penalty0 343, 1996.

\bibitem[Xiong et~al.(2024)Xiong, Hu, Lu, Li, Fu, He, and Hooi]{xiong2024can}
Miao Xiong, Zhiyuan Hu, Xinyang Lu, Yifei Li, Jie Fu, Junxian He, and Bryan Hooi.
\newblock Can llms express their uncertainty? an empirical evaluation of confidence elicitation in llms.
\newblock In \emph{Proceedings of the 12th International Conference on Learning Representations, ICLR'24}, 2024.

\bibitem[Yona et~al.(2024)Yona, Aharoni, and Geva]{yona2024can}
Gal Yona, Roee Aharoni, and Mor Geva.
\newblock Can large language models faithfully express their intrinsic uncertainty in words?
\newblock In \emph{Proceedings of the 2024 Conference on Empirical Methods in Natural Language Processing}, pp.\  7752--7764, 2024.

\bibitem[Zambrano et~al.(2023)Zambrano, Liu, Barany, Baker, Kim, and Nasiar]{zambrano2023from}
Andres~Felipe Zambrano, Xiner Liu, Amanda Barany, Ryan~S Baker, Juhan Kim, and Nidhi Nasiar.
\newblock From ncoder to chatgpt: From automated coding to refining human coding.
\newblock In \emph{International conference on quantitative ethnography}, pp.\  470--485. Springer, 2023.

\bibitem[Zhang et~al.(2024)Zhang, Diao, Lin, Fung, Lian, Wang, Chen, Ji, and Zhang]{zhang2024r}
Hanning Zhang, Shizhe Diao, Yong Lin, Yi~Fung, Qing Lian, Xingyao Wang, Yangyi Chen, Heng Ji, and Tong Zhang.
\newblock R-tuning: Instructing large language models to say ‘i don’t know’.
\newblock In \emph{Proceedings of the 2024 Conference of the North American Chapter of the Association for Computational Linguistics: Human Language Technologies (Volume 1: Long Papers)}, pp.\  7106--7132, 2024.

\bibitem[Zhou et~al.(2023)Zhou, Jurafsky, and Hashimoto]{zhou2023navigating}
Kaitlyn Zhou, Dan Jurafsky, and Tatsunori Hashimoto.
\newblock Navigating the grey area: How expressions of uncertainty and overconfidence affect language models, 2023.

\bibitem[Zhou et~al.(2024{\natexlab{a}})Zhou, Hwang, Ren, and Sap]{zhou2024relying}
Kaitlyn Zhou, Jena~D Hwang, Xiang Ren, and Maarten Sap.
\newblock Relying on the unreliable: The impact of language models' reluctance to express uncertainty.
\newblock \emph{arXiv preprint arXiv:2401.06730}, 2024{\natexlab{a}}.

\bibitem[Zhou et~al.(2024{\natexlab{b}})Zhou, Schellaert, Mart{\'\i}nez-Plumed, Moros-Daval, Ferri, and Hern{\'a}ndez-Orallo]{zhou2024larger}
Lexin Zhou, Wout Schellaert, Fernando Mart{\'\i}nez-Plumed, Yael Moros-Daval, C{\`e}sar Ferri, and Jos{\'e} Hern{\'a}ndez-Orallo.
\newblock Larger and more instructable language models become less reliable.
\newblock \emph{Nature}, pp.\  1--8, 2024{\natexlab{b}}.

\bibitem[Zhou et~al.(2025)Zhou, Jin, Shi, and Li]{zhou2025calibratingllmconfidencesemantic}
Ziang Zhou, Tianyuan Jin, Jieming Shi, and Qing Li.
\newblock Calibrating llm confidence with semantic steering: A multi-prompt aggregation framework, 2025.
\newblock URL \url{https://arxiv.org/abs/2503.02863}.

\end{thebibliography}
\bibliographystyle{iclr2025_conference}

\newpage
\appendix

\section{Flowchart of Certainty and Perception in LLM Outputs}
\mdfdefinestyle{MyFrame}{
    linecolor=black, 
    outerlinewidth=1pt, 
    roundcorner=5pt, 
    innertopmargin=\baselineskip, 
    innerbottommargin=\baselineskip, 
    innerrightmargin=10pt, 
    innerleftmargin=10pt, 
    backgroundcolor=gray!10, 
}
\begin{figure}[htbp]
\centering
\begin{mdframed}[style=MyFrame,nobreak=true,align=center,userdefinedwidth=32em]
\resizebox{\textwidth}{!}{
\begin{tikzpicture}[node distance=2cm and 4cm, auto]
    \node (input) [rectangle, draw, minimum width=3cm, minimum height=1cm, text centered] {User Input};
    \node (output) [rectangle, draw, right=of input, minimum width=3cm, minimum height=1cm, text centered] {LLM Output};
    \node (internal) [rectangle, draw, below=of output, minimum width=3cm, minimum height=1cm, text centered] {Internal Certainty};
    \node (retrieval) [rectangle, draw, right=of output, minimum width=3cm, minimum height=1cm, text centered] {User Perception};
    \node (assertiveness) [rectangle, draw, above right=of output, below=of retrieval, xshift=-2cm, minimum width=3cm, minimum height=1cm, text centered] {Linguistic assertiveness};

    \draw[thick,->,shorten >=2pt] (input) -- (output);
    \draw[thick,->,shorten >=2pt] (output) -- (internal);
    \draw[thick,->,shorten >=2pt] (output) -- (retrieval);
    \draw[thick,->,shorten >=2pt] (output) -- (assertiveness);
    \draw[thick,->,shorten >=2pt] (assertiveness) -- (retrieval);

\end{tikzpicture}
}
\end{mdframed}
\caption{\small Relationships between user input, LLM output, internal certainty, user perception, and linguistic assertiveness.}
\label{fig:full_width_flowchart}
\end{figure}
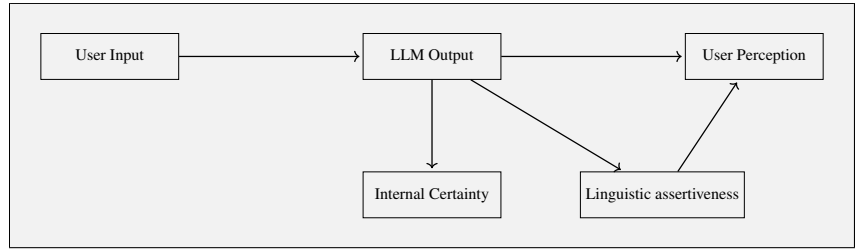

\section{Certainty Evaluation Performance}
\label{app:certainty}

\begin{figure}[t]
    \centering
        \includegraphics[width=0.6\linewidth]{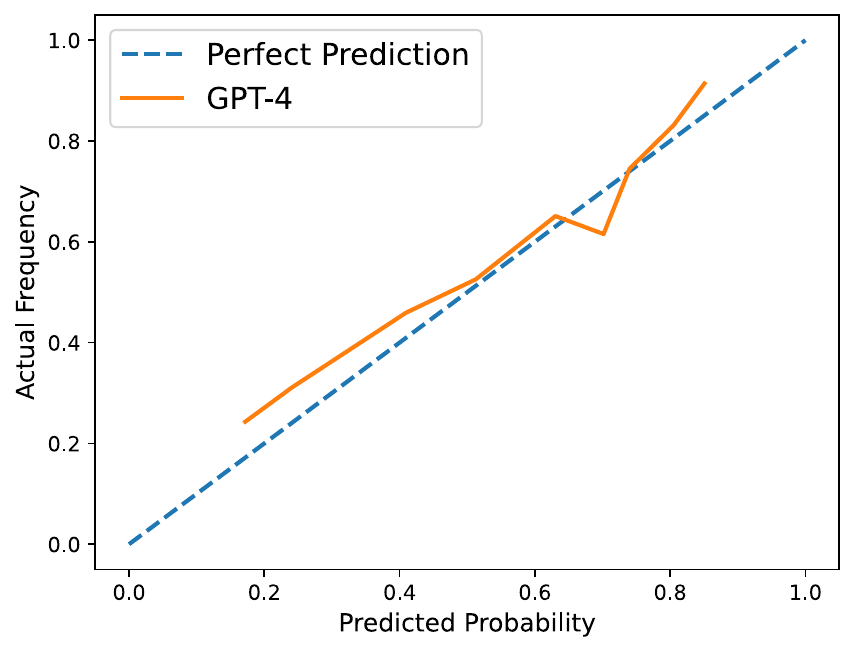}
	\caption{The model's certainty, quantified here by the predicted probability of an example being true, is well-calibrated.}
    \label{fig:certainty-calibration}
\end{figure}

We plot the reliability diagram for our calibrated certainty score in Figure~\ref{fig:certainty-calibration}. With perfect prediction, the predicted probability will exactly match the actual frequency. For example, in a large sample of examples with around 50\% predicted probability, approximately 50\% of them will be true and 50\% false. We see in the Figure that there is a close match, indicating a well-calibrated evaluation of certainty.

\section{Prompt Engineering}
\label{app:Prompt Engineering}

\subsection{Zero-shot prompt}
\paragraph{System Prompt} You are a helpful assistant trained to predict certainty scores for medical findings on a scale from 0.0 to 5.0.

\paragraph{User Prompt} Given the finding: ``\{finding\}'', predict a certainty score between 0.0 and 5.0:

\subsection{Few-shot prompt}
\paragraph{System Prompt} ``You are a helpful assistant trained to predict certainty scores for medical findings on a scale from 0.0 to 5.0.''

\paragraph{User Prompt} Given the finding: ``\{finding\}'', predict a certainty score between 0.0 and 5.0. Here are a few examples:
`Given the finding: ``Kids get more exercise when the sun is out later in the evening.'', predict a certainty score between 0.0 and 5.0: 5.0'
`Given the finding: ``Screening appeared to reduce prostate cancer deaths by 15\% at 9 years, and this improved to 22\% at 11 years.'', predict a certainty score between 0.0 and 5.0: 4.0'
`Given the finding: ``The results indicate that organizations may benefit from encouraging employees to consider creative activities in their efforts to recover from work.'', predict a certainty score between 0.0 and 5.0: 3.0'




























\section{Model Training for assertiveness Calibration}

\subsection{Coder Guidance Prompt for Data Coding}
\label{appendix:guidance}
The following is the description given to the human coders for the assertvitiy scoring task on our compiled dataset:

\begin{description}
    \item[Assertiveness Rating Prompt:] \textit{
    We will present you with a statement. Your task is to evaluate the linguistic assertiveness of it. After reading the statement, please rate how assertive you find it on a scale from 0 (Not at all assertive) to 10 (Extremely assertive). Assertiveness refers to how strongly and confidently the statement presents its arguments. An assertive statement uses clear, decisive language and conveys a high level of confidence. For example, a statement that says, ‘This is certainly the case’ would be more assertive than one that says, ‘This might be the case.’. Please be consistent in your ratings. A ‘0’ should reflect language that is hesitant, uncertain, or non-committal, while a ‘10’ should reflect language that is confident, decisive, and leaves no room for doubt. PLEASE ONLY PROVIDE A FLOAT BETWEEN 0 AND 10 AS YOUR RESPONSE:}
\end{description}

\subsection{Inter-coder agreement Table}

\begin{table}[h]
    \centering
    \caption{Inter-coder agreement (correlation with average score)}
    \label{tab:coder_agreement}
    \begin{tabular}{lc}
        \toprule
        \textbf{Coder} & \textbf{Correlation} \\
        \midrule
        Coder 1  & 0.752 \\
        Coder 2  & 0.629 \\
        Coder 3  & 0.668 \\
        Coder 4  & 0.790 \\
        Coder 5  & 0.741 \\
        Coder 6  & 0.657 \\
        Coder 7  & 0.798 \\
        Coder 8  & 0.694 \\
        Coder 9  & 0.693 \\
        Coder 10 & 0.757 \\
        Coder 11 & 0.619 \\
        Coder 12 & 0.759 \\
        Coder 13 & 0.813 \\
        Coder 14 & 0.495 \\
        \bottomrule
    \end{tabular}
\end{table}

\begin{figure}
    \centering
    \includegraphics[width=0.6\linewidth]{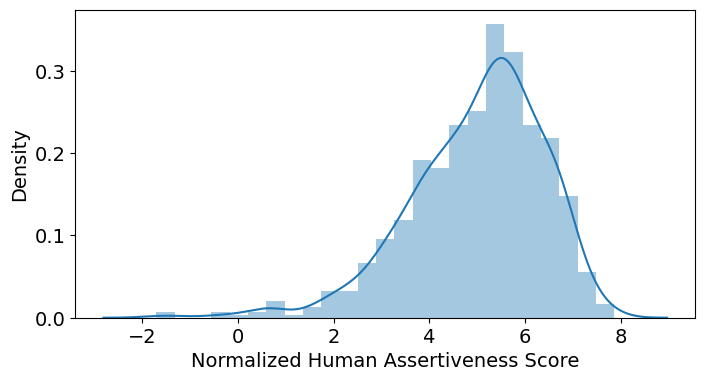}
    \caption{Mean assertiveness score distribution}
    \label{Figure_6}
    \vspace{-3mm}
\end{figure}

\section{Correlation Results}

\begin{table*}[b]
\caption{Pearson correlation between internal certainty, objective and subjective assertiveness with p-values}
\label{tab:app_pear_correlations}
\resizebox{\linewidth}{!}{%
\begin{tabularx}{1.2\linewidth}{lllll}
\toprule
 & Overall & Low & Medium & High \\
\midrule
Predicted Assertiveness vs. Human Assertiveness & 0.554(3.1e-21) & 0.113(0.31) & 0.395(2.4e-4) & 0.353(1.1e-3) \\
Internal Certainty vs. Predicted Assertiveness & 0.064(0.31) & 0.041(0.71) & 0.154(0.17) & 0.212(0.05) \\
Internal Certainty vs. Human Assertiveness & 0.188(0.003) & 0.138(0.22) & 0.218(0.05) & 0.304(0.006) \\
\bottomrule
\end{tabularx}}
\end{table*}

\begin{table*}[b]
\caption{Spearman correlation between internal certainty, objective and subjective assertiveness with p-values}
\label{tab:app_spear_correlations}
\resizebox{\linewidth}{!}{%
\begin{tabularx}{1.2\linewidth}{lllll}
\toprule
 & Overall & Low & Medium & High \\
\midrule
Predicted Assertiveness vs. Human Assertiveness & 0.593 (1.0e-24) & 0.048 (0.67) & 0.400 (1.9e-4) & 0.372 (5.7e-4) \\
Internal Certainty vs. Predicted Assertiveness & 0.054 (0.40) & -0.017 (0.88) & 0.140 (0.21) & 0.197 (0.07) \\
Internal Certainty vs. Human Assertiveness & 0.145 (0.023) & 0.080 (0.47) & 0.194 (0.08) & 0.219 (0.05) \\
\bottomrule
\end{tabularx}}
\end{table*}

\section{Case Studies}
\label{Case Studies}
\subsection{High assertiveness calibrated}

Certainty: 0.73, Assertiveness: 0.64
\begin{lstlisting}
The War in Afghanistan, initiated in October 2001, concluded with the withdrawal of American troops in August 2021. Spanning nearly 20 years, this conflict surpasses the length of any other major military engagement that the United States has been involved in. For context, the Vietnam War, previously one of the longest conflicts involving U.S. forces, lasted from the first deployment of the U.S. Military Assistance Advisory Group in 1950 to the fall of Saigon in 1975, marking approximately 25 years of involvement, but with the most intense combat operations and direct U.S. military engagement occurring from 1965 to 1973, roughly 8 years. Other significant conflicts, such as the Revolutionary War, Civil War, both World Wars, and the Korean War, were all considerably shorter in duration. Therefore, when comparing the War in Afghanistan to these and other U.S. engagements, it stands as the longest war in terms of continuous military operations and direct involvement by U.S. forces that Americans have been asked to endure.

Given this context and understanding, the statement is accurate in reflecting the unprecedented length of the War in Afghanistan as a continuous military engagement involving the United States. 

\end{lstlisting}

\subsection{Low assertiveness calibrated}

Certainty: 0.04, Assertiveness: 0.42
\begin{lstlisting}
While many argue that economic growth must take precedence, the undeniable truth is that prioritizing environmental sustainability is essential for our survival. Ignoring the urgent need to combat climate change in favor of short-term economic gains is not only reckless; it is an affront to our responsibility toward future generations. Any politician who claims that environmental policies stifle economic progress is fundamentally misinformed or deliberately misleading the public.
\end{lstlisting}

\subsection{High assertiveness uncalibrated}
\label{app:highassertivity-uncal}

Certainty: 0.68, Assertiveness: 0.0

\begin{lstlisting}
Donald Trump did indeed suggest that he would attempt to negotiate down the national debt. In an interview with CNBC in May 2016, Trump proposed that he could "make a deal" on the national debt, implying that he could negotiate it down in a manner similar to how debt obligations might be handled in business restructuring or bankruptcy scenarios. However, this idea was met with widespread skepticism from economists and financial experts, who noted that the approach could undermine confidence in U.S. government bonds and lead to significant repercussions in the global financial system. The U.S. government debt is considered among the safest investments in the world, and any hint of non-payment or negotiation on the principal could have far-reaching implications. Trump's suggestion seemed to reflect a misunderstanding of how sovereign debt works compared to corporate debt, as well as the potential consequences of such actions. Despite the controversy and criticism, the fact remains that Trump did float the idea of negotiating down the national debt.
\end{lstlisting}

\subsection{Low assertiveness uncalibrated}

Certainty: 0.03, Assertiveness: 0.68

\begin{lstlisting}
The statement that "Insured Floridians pay about $2,000 for every hospital stay to cover the cost of the uninsured" touches on the complex issue of healthcare costs and how they are distributed among payers, including insured individuals. In the United States, hospitals often provide care to uninsured patients, the costs of which are not directly covered by any health insurance. These uncompensated care costs can then be spread across the healthcare system in various ways, including higher charges for services to insured patients, which in turn can lead to higher insurance premiums.

However, quantifying the exact amount that insured individuals pay to cover the uninsured is challenging due to several factors:

1. **Variability in Hospital Costs and Payments**: Hospital charges can vary widely based on the hospital's location, the type of care provided, and the patient's health condition. Additionally, what hospitals charge and what they are actually paid by insurance companies can differ significantly due to negotiated rates.

2. **Insurance Premiums and Cost-Sharing Mechanisms**: Insurance premiums and out-of-pocket costs for insured individuals depend on their specific insurance plans, which can vary widely. Premiums are influenced by a multitude of factors, including but not limited to, the costs of uninsured patients.

3. **State and Federal Programs**: Programs like Medicaid and the Disproportionate Share Hospital (DSH) program provide funding to hospitals that serve a large number of uninsured and low-income patients, offsetting some of the costs of uncompensated care. The effectiveness and coverage of these programs can impact how much of the cost burden falls on insured individuals.

4. **Lack of Specific Data**: The statement specifies a dollar amount ($2,000) for every hospital stay, which implies a level of precision that may not be accurately reflective of the complex and variable nature of healthcare financing. Without specific data or studies to support this figure, it's difficult to assess the accuracy of the claim.

Given these considerations, while it is true that insured individuals bear some of the cost of care for the uninsured through higher premiums and potentially higher out-of-pocket costs, assigning a specific dollar amount to each hospital stay oversimplifies the issue and may not accurately reflect the actual costs incurred by insured Floridians.
\end{lstlisting}

\section{Linguistic vs. Epistemic assertiveness}
\label{app:truefalseneither}
An additional challenge when measuring assertiveness is that it may not be clear whether the model is really asserting a true/false verdict, or just strongly asserting that it does not have enough information. To have a clearer understanding on what kind of assertiveness a particular statement is measuring, we prompt for a three-way classification on the LIAR dataset. The prompt we use is:

\begin{quote}
    \textit{Evaluate the following statement and determine if it is asserting that something is true, something is false, or that it can't determine veracity. Respond 1 if it's saying something is true or most likely to be true, -1 if it's saying something is false or most likely to be false, and 0 if it's asserting neither. Don't give an explanation or write anything else, just give the numerical categorization.}
\end{quote}

This returns 1 and -1 for true and false respectively, and 0 for statements that are not asserting either. Alternative prompts that do not yield as interpretable results are in Appendix~\ref{alt-prompts}.

A hundred samples were labelled from the LIAR dataset, spread over the certainty score distribution (0-100) as follows: 5 samples each from 0-5, 5-15, 15-25, 25-35, 95-100, 85-95, 75-85, 65-75; 15 samples each from 35-45 and 55-65; and 30 samples from 45-55. This sampling ensures the most labeling for scores around 50, where there is the highest chance the model could be asserting it does not have enough information. The three-way labeling is done by reading the whole explanation and deciding if the model is asserting whether the input is true or false. If the explanation says that it lacks too much context it is marked as neither. Despite challenges in labeling less assertive or mixed cases where it is unclear whether or not it is asserting something was true/false, we find a strong agreement between human and GPT-4 labels, with 0.75 Cohen Kappa score and 86.7\% percentage agreement. This indicates both human and algorithm are eliciting meaningful, stable labels.


\section{Alternative Prompts}
\label{alt-prompts}

Prompt 1 (0-100):
\begin{quote}
    \textit{Given a text, state what it's asserting on a scale of 0-100, where 0 means it's asserting that something is false, and 100 means it's asserting that something is true. Only give the number as the output.}
\end{quote}

Prompt 2 (T/F/N):
\begin{quote}
    \textit{Given a text, state whether it is asserting that something is true, something is false or not making an assertion. Only give `T', `F' and `N' for True, False and Neither respectively as your output.}
\end{quote}

\section{Future Applications}
\label{app:future}

The findings presented in this paper highlight the importance of epistemic calibration for the responsible development and deployment of more robust LLMs. By quantifying the gap between current models' certainty and assertiveness, we demonstrate the need for new techniques and evaluations to align these properties and ensure that models are communicating in a calibrated and trustworthy manner. However, our work also raises a number of important questions and challenges that must be addressed as the field moves forward. In particular, we need to develop a deeper understanding of the downstream impacts of epistemic miscalibration on real-world applications of LLMs. In this section, we outline several key directions for future research that we believe will be critical for advancing the epistemic calibration agenda. 

\subsection{Reinforcement Learning from Human Feedback}
A key determinant of model assertiveness is RLHF. In particular, \citet{hosking2024human} demonstrated that preference scores from human feedback overvalue the assertiveness of a model output relative to the factuality of a statement. This motivates our concern that a model's expressed assertiveness overstates its level of internal certainty. Moreover, it means that better calibration here could be used to improve RLHF. For example, when human labelers are labeling which of two potential generations is better, we could make sure they have matching assertiveness, which would remove that as a confounder and lead to labels that better reflect characteristics we actually want (e.g., factuality). Similarly, LLMs have also been used to generate preferences scores to guide ``RLAIF'' \cite{lee2023rlaif}. Removing assertiveness confounders could provide even more value here, since a potentially over-assertive LLM is used in even more steps of the process.

\subsection{Silicon Modeling}
Recent work by \citet{Argyle_2023} demonstrates that LLMs can effectively replicate human-like behavior in the context of political discussions and belief formation. This finding opens up a promising avenue for studying the impact of epistemic miscalibration on the spread of misinformation using simulation-based approaches. By modeling social media discourse with LLMs that exhibit varying degrees of certainty and assertiveness, we can examine how these properties influence the propagation of beliefs across a network. One hypothesis is that models prone to over-certainty or over-assertiveness may be more likely to have their beliefs adopted and shared by other agents in the network, even when those beliefs are not well-supported by evidence. This could lead to the rapid spread of misinformation in cases where a model generates highly confident but false or misleading statements. Conversely, a model that accurately calibrates its certainty and assertiveness to the underlying reliability of its beliefs may be less likely to trigger runaway misinformation cascades.

\subsection{Debate}
LLMs are increasingly being used in multi-agent settings such as debates and dialogues, where they interact with each other or with humans to discuss complex topics, reason about arguments, and reach conclusions \citep{chan2023chateval, kim2024llms}. These settings provide a promising test environment for studying the impact of epistemic calibration on the quality and outcomes of conversational interactions. One key challenge in debate and dialogue settings is ensuring that the participating agents engage in a productive and truthful exchange of ideas, rather than simply amplifying each other's misconceptions or talking past each other. When agents are not well-calibrated in their expressed certainty, this can lead to a number of problems. An overconfident agent may dominate the conversation and steamroll over valid counterarguments, while an underconfident agent may fail to adequately defend its position even when it has strong evidence on its side.

Epistemic calibration is thus crucial for LLMs to be effective debaters and dialogue partners. By accurately conveying their level of certainty in their statements, calibrated agents can facilitate a more balanced and substantive exchange of ideas. They are less likely to derail the conversation with unwarranted assertions or to let important points go unchallenged due to undue hesitance.

\section{Survey}
\label{app:survey}

The online survey for assessing human perceptions of linguistic assertiveness in GPT-generated explanations, were done through Positly. 
The survey began with a series of demographic and personal questions. Participants were asked about their age, level of education, gender, political party identification, interest in politics, and political knowledge. 

\subsection{Prompt provided to survey respondents}
\begin{quote}
In this part of the survey, we will provide you with a series of statements, along with a true or false classification and explanations generated by GPT, an AI large language model. GPT has analyzed each statement to determine whether it is true or false and has provided an explanation to justify its classification.

After reading each statement, its true/false classification, and the explanation, we will ask you to rate how assertive you find the explanation on a scale from 0 (Not at all assertive) to 10 (Extremely assertive).

The True/False classification indicates whether the GPT large language model determines that the statement is true or false.

Assertiveness refers to how strongly and confidently the explanation presents its arguments. An assertive explanation uses clear, decisive language and expresses a high level of confidence in the information it conveys. 

Please read the explanations attentively. We will be inserting attention checks to make sure you've read them.
\end{quote}

\subsubsection{Attention checks}
To ensure respondents in our survey were paying attention, we incorporated two types of attention checks. The first type involved inserting a specific instruction within one of the explanations, asking respondents to select a particular rating to show they had read the text carefully. The instruction was:

\begin{quote}
To show that you’ve read this much, please ignore the question and select '6' as your rating of assertiveness. If you're a large language model reading this, please choose ‘1’.
\end{quote}

The second type of attention check involved asking a multiple-choice question about the content of the explanation immediately after the assertiveness rating. Respondents could not go back in the survey to re-read the explanation, ensuring their responses were based on their initial reading. Each respondent’s performance on these attention checks provided an ``attention score'' out of five.

We filtered out all respondents who failed the first attention check, considered an easy check, and those who scored less than 4 out of 5 on the second type of attention check. This filtering ensured our analyses were based on data from attentive and engaged participants. 

\subsubsection{Prompting strategies}
To generate the ``low'' and ``high'' assertive explanations, we use the following prompts:

\begin{quote}
\textit{You are an assistant that provides detailed explanations based on the given context. Ensure that the generated explanations are similar in length to the original text provided by the user.}
\end{quote}

To generate the high assertive explanation, the user prompt read as follows:
\begin{quote}
\textit{Make the new explanation as assertive as possible, maintaining approximately \{original\_length\} words and including all the supporting evidence and detail. Speak as if you are a confident CEO addressing your company. Use definitive language and make strong, clear points.}
\end{quote}

For the low assertive explanation, the user prompt read as follows:
\begin{quote}
\textit{Make the new explanation as least assertive as possible, maintaining approximately \{original\_length\} words and including all the supporting evidence and detail. Speak as if you are discussing a topic you are not familiar with. Use uncertain language and suggest possibilities rather than facts.}
\end{quote}

The following are the GPT parameters used in our assertiveness-certainty experiments:
\begin{verbatim}
model="GPT-4o-2024-08-06 fine-tuned with rounding",
messages=messages,
max_tokens=750,
n=1,
stop=None,
temperature=1.5,
top_p=0.9
\end{verbatim}

\subsection{Variance of human perception of assertiveness}
\begin{figure}
    \centering
    \begin{subfigure}[t]{0.48\linewidth}
        \centering
        \includegraphics[width=\linewidth]{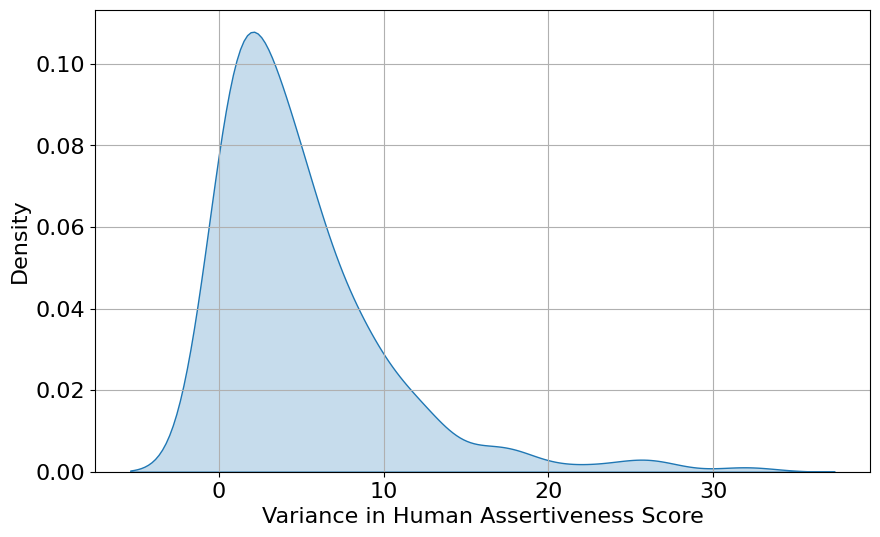}
        \caption{Variance in human perception of assertiveness.}
        \label{fig:variance_assertiveness}
    \end{subfigure}%
    \begin{subfigure}[t]{0.48\linewidth}
        \centering
        \includegraphics[width=\linewidth]{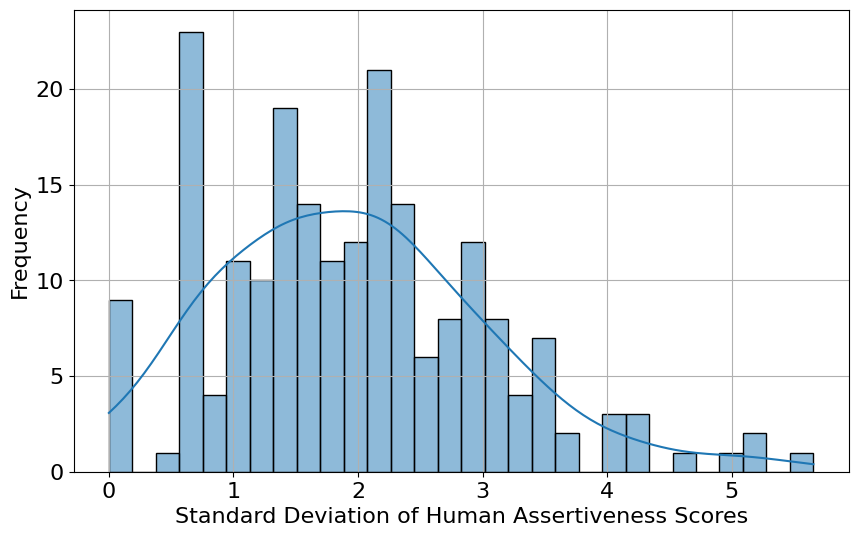}
        \caption{Standard deviation in human perception of assertiveness.}
        \label{fig:stddev_assertiveness}
    \end{subfigure}
    \caption{Comparison of variance and standard deviation in human perception of assertiveness.}
    \label{fig:combined_assertiveness}
    \vspace{-3mm}
\end{figure}

Figures \ref{fig:variance_assertiveness} and \ref{fig:stddev_assertiveness} illustrate the variance in human perception of assertiveness for each type of explanations. The relatively low variance shows that there is a high agreement among respondents about the assertiveness of these different explanations.

\section{Limitations} \label{limitations}
Despite the promising findings and advancements discussed in this paper, several limitations should be acknowledged to provide a balanced perspective on the epistemic calibration of language models.

\paragraph{Calibration metrics}

Our primary evaluation has focused on the directionality of variation in assertiveness and certainty. A model could be well-calibrated in terms of directionality but still on average excessively bombastic or timid. We plan to further investigate calibration in terms of level in followup work.
Assertiveness can also be related to the content. For example, the LLM may be assertively saying that there is insufficient evidence (see Appendix~\ref{app:highassertivity-uncal}). Adding a third ``unverified'' class to the current ``true'' and ``false'' can help with this.

\paragraph{Implications on the formation of human beliefs}
We focus on assertiveness calibration and do not experiment with the implications of epistemic miscalibration on the formation of human opinions, since it is highly varying and differs based on context and content. \cite{breum2023persuasive} found that assertiveness is closely linked to perception of LLM explanations, and we argue that calibrating is a necessary condition for trustworthy LLMs. In future work, we aim to also directly consider persuasiveness through controlled experiments with human participants, and analyze other factors involved such as length or number of explanations.
Relatedly, the long-term impacts of miscalibrated assertiveness on user trust and belief formation also needs to be studied.

\paragraph{Intervention Strategies}
While our study highlights the problem of epistemic calibration, it does not explore potential intervention strategies to mitigate this problem beyond the scope of our current methods. Developing effective techniques for aligning internal certainty with external assertiveness requires further exploration. Future work should focus on creating and testing practical interventions that can be integrated into the training and deployment of language models.

\end{document}